\documentclass{article}

\PassOptionsToPackage{numbers, compress}{natbib}
\usepackage[final]{neurips_2020}

\usepackage[utf8]{inputenc} %
\usepackage[T1]{fontenc}    %
\usepackage{hyperref}       %
\usepackage{url}            %
\usepackage{booktabs}       %
\usepackage{amsfonts}       %
\usepackage{nicefrac}       %
\usepackage{microtype}      %
\usepackage{bm}

\usepackage{natbib}

\usepackage[noend]{algorithmic}
\usepackage{algorithm}
\newcommand{\beeq}{\begin{equation}}
\newcommand{\eneq}{\end{equation}}
\usepackage{amsmath,amssymb}
\DeclareMathOperator*{\argmax}{arg\,max}
\DeclareMathOperator*{\argmin}{arg\,min}
\usepackage{graphicx}
\usepackage{wrapfig}
\usepackage[normalem]{ulem}
\usepackage{multirow}

\makeatletter
\newcommand{\figcaption}[1]{\def\@captype{figure}\caption{#1}}
\newcommand{\tblcaption}[1]{\def\@captype{table}\caption{#1}}
\newcommand{\bmx}{\bm{\mathrm{x}}}
\newcommand{\bmz}{\bm{\mathrm{z}}}
\newcommand{\bmw}{\bm{\mathrm{w}}}
\newcommand{\bmq}{\bm{\mathrm{q}}}
\newcommand{\bmf}{\bm{\mathrm{f}}}
\newcommand{\bmv}{\bm{\mathrm{v}}}
\newcommand{\bmg}{\bm{\mathrm{g}}}
\newcommand{\bmdelta}{\bm{\delta}}
\makeatother

\title{Diversity Can Be Transferred: Output Diversification for White- and Black-box Attacks}

\author{%
  Yusuke Tashiro\textsuperscript{123*}, Yang Song\textsuperscript{1}, Stefano Ermon\textsuperscript{1}\\
  \textsuperscript{1}Department of Computer Science, Stanford University, Stanford, CA, USA\\
  \textsuperscript{2}Mitsubishi UFJ Trust Investment Technology Institute, Tokyo, Japan\\
  \textsuperscript{3}Japan Digital Design, Tokyo, Japan\\
  \texttt{\{ytashiro,songyang,ermon\}@stanford.edu}
}

\begin{document}

\maketitle

\begin{abstract}

Adversarial attacks often involve random perturbations of the inputs drawn from uniform or Gaussian distributions, e.g., to initialize optimization-based white-box attacks or generate update directions in black-box attacks.
These simple perturbations, however, could be sub-optimal as they are agnostic to the model being attacked.
To improve the efficiency of these attacks, we propose Output Diversified Sampling (ODS), a novel sampling strategy that attempts to maximize diversity in the target model's outputs among the generated samples. 
While ODS is a gradient-based strategy, the diversity offered by ODS is transferable and can be helpful for both white-box and black-box attacks via surrogate models.
Empirically, we demonstrate that ODS significantly improves the performance of existing white-box and black-box attacks. 
In particular, ODS reduces the number of queries needed for state-of-the-art black-box attacks on ImageNet by a factor of two.
\end{abstract}

\section{Introduction}
Deep neural networks have achieved great success in image classification. However, it is known that they are vulnerable to adversarial examples~\citep{Szegedy13} --- small perturbations imperceptible to humans that cause classifiers to output wrong predictions. 
Several studies have focused on improving model robustness against these malicious perturbations. Examples include adversarial training~\citep{madry17,Ian15}, %
input purification using generative models~\citep{song2017pixeldefend,samangouei2018defense}, 
regularization of the training loss~\citep{regGrad18,regStGrad18,regCurv19,regLinear19}, 
and certified defenses~\citep{Certify17,Certify18_2,Certify19}.

Strong attacking methods are crucial for evaluating the robustness of classifiers and defense mechanisms. 
Many existing adversarial attacks rely on random sampling, i.e., adding small random noise to the input.
In white-box settings, random sampling is widely used for random restarts~\citep{PGD17,Dist19,fab19,MT19} to find a diverse set of starting points for the attacks.
Some black-box attack methods also use random sampling to explore update directions~\citep{Brendel18,Guo19} or to estimate gradients of the target models~\citep{ilyas2018blackbox,IEM2018PriorCB,autozoom2019}. 
In these attacks, random perturbations are typically sampled from a na\"{i}ve uniform or Gaussian distribution in the input pixel space.

\begin{figure*}[htbp]
\centering
    \begin{tabular}{ccc|ccc}

      \begin{minipage}{0.2\hsize}
        \begin{center}
          \includegraphics[width=0.88\textwidth]{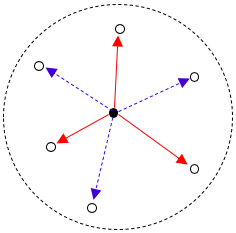}
        \end{center}
      \end{minipage}
        &
      \begin{minipage}{0.2\hsize}
        \begin{center}
          \includegraphics[width=0.908\textwidth]{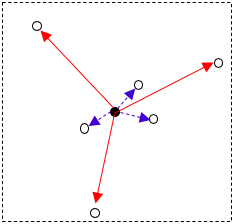}
        \end{center}
      \end{minipage} &&&
      \begin{minipage}{0.2\hsize}
        \begin{center}
          \includegraphics[width=0.908\textwidth]{image/output_target.png}
        \end{center}
      \end{minipage} &
      \begin{minipage}{0.2\hsize}
        \begin{center}
          \includegraphics[width=0.908\textwidth]{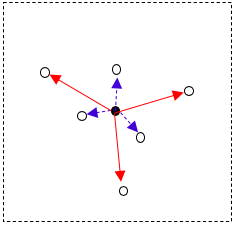}
        \end{center}
      \end{minipage}
      \\ 
     Input space& Output space&&& Output space &Output space 
      \\
     &&&& (surrogate model) &(target model)
    \end{tabular}
    \caption{Illustration of the differences between random sampling (blue dashed arrows) and ODS (red solid arrows). In each figure, the black `o' corresponds to an original image, and white `o's represent sampled perturbations. (Left): white-box setting. Perturbations by ODS in the input space are crafted by maximizing the distance in the output space. (Right): black-box setting. Perturbations crafted on the surrogate model transfer well to perturbations on the target model. 
    }
    \label{figure1}
    \vskip -0.0in
\end{figure*}

Random sampling in the input space, however, may not sufficiently explore the output (logits) space of a neural network --- diversity in the input space does not directly translate to diversity in the output space of a deep nonlinear model. We illustrate this phenomenon in the left panel of Figure~\ref{figure1}. When we add random perturbations to an image in the input space (see dashed blue arrows in the first plot of Figure~\ref{figure1}), the corresponding output logits could be very similar to the output for the original image (as illustrated by the second plot of Figure~\ref{figure1}). 
Empirically, we observe that this phenomenon can negatively impact the performance of attack methods.

To overcome this issue, we propose a sampling strategy designed to obtain samples that are diverse in the output space. Our idea is to perturb an input away from the original one as measured directly by distances in the output space (see solid red arrows in the second plot in Figure~\ref{figure1}). 
First, we randomly specify a direction in the output space. Next, we perform gradient-based optimization to generate a perturbation in the input space that 
yields a large change in the specified direction. 
We call this new sampling technique \uline{Output Diversified Sampling} (ODS).

ODS can improve adversarial attacks under both white-box and black-box settings. For white-box attacks, we exploit ODS to initialize the optimization procedure of finding adversarial examples (called ODI). ODI typically provides much more diverse (and effective) starting points for adversarial attacks. 
Moreover, this initialization strategy is agnostic to the underlying attack method, and can be incorporated into most optimization-based white-box attack methods.
Empirically, we demonstrate that ODI improves the performance of $\ell_\infty$ and $\ell_2$ attacks compared to na\"{i}ve initialization methods.
In particular, the PGD attack with ODI outperforms the state-of-the-art MultiTargeted attack~\citep{MT19} against pre-trained defense models, with 50 times smaller computational complexity on CIFAR-10.

In black-box settings, we cannot directly apply ODS because we do not have access to gradients of the target model. As an alternative, we apply ODS to surrogate models and observe that the resulting samples are diverse with respect to the target model: diversity in the output space transfers (see the rightmost two plots in Figure~\ref{figure1}). 
Empirically, 
we demonstrate that ODS can reduce the number of queries needed for a score-based attack (SimBA~\citep{Guo19}) by a factor of two on ImageNet. 
{ODS also shows better query-efficiency than P-RGF~\citep{Cheng19prior}, which is another method exploiting surrogate models to improve a black-box attack.}
These attacks with ODS achieve better query-efficiency than the state-of-the-art Square Attack~\citep{ACFH2019square}.
In addition, ODS with a decision-based attack (Boundary Attack~\citep{Brendel18}) reduces the median perturbation distances of adversarial examples by a factor of three compared to the state-of-the-art HopSkipJump~\citep{chen2019hop} and Sign-OPT~\citep{cheng20sign} attacks.

\section{Preliminaries}
We denote an image classifier as $\bmf: \bmx\in[0,1]^D \mapsto \bmz\in \mathbb{R}^{C}$, where $\bmx$ is an input image, $\bmz$ represents the logits, and $C$ is the number of classes.
We use
$ h(\bmx) = \argmax_{c=1,\ldots,C} f_{c}(\bmx)$ to denote the model prediction, where $f_{c}(\bmx)$ is the $c$-th element of $\bmf(\bmx)$. %

Adversarial attacks can be classified into targeted and untargeted attacks.
Given an image $\bmx$, a label $y$ and a classifier $\bmf$, 
the purpose of untargeted attacks  
is to find an adversarial example $\bmx^{\text{adv}}$ that is similar to $\bmx$ but causes misclassification $ h(\bmx^{\text{adv}}) \neq y $. 
In targeted settings, attackers aim to change the model prediction $h(\bmx^{\text{adv}})$ to a particular target label $t \neq y$. 
The typical goal of adversarial attacks is to find an adversarial example $\bmx^{\text{adv}}$ within $B_{\epsilon}(\bmx) = \{ \bmx + \bmdelta : \|\bmdelta\|_{p} \leq \epsilon \}$, i.e., the $\epsilon$-radius ball around an original image $\bmx$. Another common setting is to find a valid adversarial example with the smallest $\ell_p$ distance from the original image. 

\paragraph{White-box attacks}
In white-box settings, attackers can access full information of the target model. One strong and popular example is the Projected Gradient Descent (PGD) attack~\citep{madry17}, which iteratively applies the following update rule:
\begin{equation}
\bmx^{\text{adv}}_{k+1}  
=  \mathrm{Proj}_{B_{\epsilon}(\bmx)} \left( \bmx^{\text{adv}}_k + \eta \, \mathrm{sign} \left(\nabla_{\bmx^{\text{adv}}_k} L(\bmf(\bmx^{\text{adv}}_k),y) 
 \right) \right)
\label{eq_pgd}
\end{equation}
where $\mathrm{Proj}_{B_{\epsilon}(\bmx)}(\bmx^{\text{adv}}) \triangleq \argmin_{\bmx' \in B_{\epsilon}(\bmx)} \|\bmx^{\text{adv}}-\bmx'\|_{p}$, $\eta$ is the step size, and $L(\bmf(\bmx),y)$ is a loss function, e.g. the margin loss defined as $\max_{i \neq y} f_{i}(\bmx) - f_{y}(\bmx)$. 
To increase the odds of success, the procedure is restarted multiple times with uniformly sampled initial inputs from $B_{\epsilon}(\bmx)$.

\paragraph{Black-box attacks}
In black-box settings, the attacker only has access to outputs of the target model without knowing its architecture and weights. Black-box attacks can be largely classified into transfer-based, score-based, and decision-based methods respectively. Transfer-based attacks craft white-box adversarial examples with respect to surrogate models, and transfer them to the target model. The surrogate models are typically trained with the same dataset as the target model so that they are close to each other.
In score-based settings, attackers can know the output scores (logits) of the classifier; while for decision-based settings, attackers can only access the output labels of the classifier. 
For these two approaches, attacks are typically evaluated in terms of query efficiency, i.e. the number of queries needed to generate an adversarial example and its perturbation size.

Recently, several studies~\citep{Cheng19prior,subspaceattack,Cai19transferSMBdirect} employed surrogate models to estimate the gradients of the loss function of the target model.
Some attack methods used random sampling in the input space, 
such as the decision-based Boundary Attack~\citep{Brendel18} and the score-based Simple Black-Box Attack~\citep{Guo19}.

\section{Output Diversified Sampling}

As intuitively presented in Figure~\ref{figure1}, random sampling in the input space does not necessarily produce samples with high diversity as measured in the output space. To address this problem, we propose Output Diversified Sampling (ODS). Given an image $\bmx$, a classifier $\bmf$ and the direction of diversification $\bmw_{\text{d}}  \in \mathbb{R}^{C}$, 
we define the normalized perturbation vector of ODS as follows:
\begin{equation}
\bmv_{\text{ODS}}(\bmx,\bmf,\bmw_{\text{d}}) = \frac{\nabla_{\bmx} (\bmw_{\text{d}}^\intercal \bmf(\bmx))}{\| \nabla_{\bmx} (\bmw_{\text{d}}^\intercal \bmf(\bmx)) \|_2},
\end{equation}
where $\bmw_{\text{d}}$ is sampled from the uniform distribution over $[-1,1]^C$. 
Below we show how to enhance white- and black-box attacks with ODS.

\subsection{Initialization with ODS for white-box attacks}
\label{sec_ODS_white}
In white-box settings, we utilize ODS for initialization (ODI) to generate output-diversified starting points. Given an original input $\bmx_{\text{org}}$ and the direction for ODS $\bmw_{\text{d}}$, we try to find a restart point $\bmx$ that is as far away from $\bmx_{\text{org}}$ as possible by maximizing $\bmw_{\text{d}}^\intercal (\bmf(\bmx)-\bmf(\bmx_{\text{org}}))$ via the following iterative update:
\begin{equation}
\label{eq_linf}
\bmx_{k+1}  = \mathrm{Proj}_{B(\bmx_{\text{org}})} \left( \bmx_{k} + \eta_{\text{ODI}} \,  \mathrm{sign}(\bmv_{\text{ODS}}(\bmx_k,\bmf,\bmw_{\text{d}})) \right)
\end{equation}
where $B(\bmx_{\text{org}})$ is the set of allowed perturbations, which is typically an $\epsilon$-ball in $\ell_p$ norm, and $\eta_{\text{ODI}}$ is a step size. When applying ODI to $\ell_2$ attacks, we omit the sign function. 
After some steps of ODI, we start an attack from the restart point obtained by ODI. 
We sample a new direction $\bmw_{\text{d}}$ for each restart in order to obtain diversified starting points for the attacks.
We provide the pseudo-code for ODI in Algorithm~\ref{alg:ap_white} of the Appendix.

One sampling step of ODI costs roughly the same time as one iteration of most gradient-based attacks (e.g., PGD). 
Empirically, we observe that the number of ODI steps $N_{\text{ODI}}=2$ is already sufficient to obtain diversified starting points (details of the sensitivity analysis are in Appendix~\ref{appendix:ap_white_sensitivity}), and fix $N_{\text{ODI}}=2$ in all our experiments unless otherwise specified. We emphasize that ODS is not limited to PGD, and can be applied to a wide family of optimization-based adversarial attacks.

\textbf{Experimental verification of increased diversity:} 
We quantitatively evaluate the diversity of starting points in terms of pairwise distances of output values $\bmf(\bmx)$, confirming the intuition presented in the left two plots of Figure~\ref{figure1}. We take a robust model on CIFAR-10 as an example of target models, and generate starting points with both ODI and standard uniform initialization to calculate the mean pairwise distance. The pairwise distance (i.e. diversity) obtained by ODI is 6.41, which is about 15 times larger than that from uniform initialization (0.38). In addition, PGD with the same steps as ODI does not generate diverse samples (pairwise distance is 0.43). 
Details are in Appendix~\ref{appendix:ap_white_diversity}.

\subsection{Sampling update directions with ODS for black-box attacks}
\label{sec_ODS_black}
In black-box settings, we employ ODS to sample update directions instead of random sampling. 
Given a target classifier $\bmf$, we cannot directly calculate the ODS perturbation  $\bmv_{\text{ODS}}(\bmx,\bmf,\bmw_{\text{d}})$ because gradients of the target model $\bmf$ are unknown. Instead, we introduce a surrogate model $\bmg$ and calculate the ODS vector $\bmv_{\text{ODS}}(\bmx,\bmg,\bmw_{\text{d}})$. 

ODS can be applied to attack methods that rely on random sampling in the input space. Since many black-box attacks use random sampling to explore update directions~\citep{Brendel18,Guo19} or estimate gradients of the target models~\citep{ilyas2018blackbox,autozoom2019,chen2019hop}, ODS has broad applications. 
In this paper, we apply ODS to two popular black-box attacks that use random sampling: decision-based Boundary Attack~\citep{Brendel18} and score-based Simple Black-Box Attack (SimBA~\citep{Guo19}). In addition, we compare ODS with P-RGF~\citep{Cheng19prior}, which is an another attack method using surrogate models.

To illustrate how we apply ODS to existing black-box attack methods, we provide the pseudo-code of SimBA~\citep{Guo19} with ODS in Algorithm~\ref{alg:black}.
The original SimBA algorithm picks an update direction $\bmq$ randomly from a group of candidates $Q$ that are orthonormal to each other. We replace it with ODS, as shown in the line 5 and 6 of Algorithm~\ref{alg:black}. For other attacks, we replace random sampling with ODS in a similar way.
Note that in Algorithm~\ref{alg:black}, we make use of multiple surrogate models and uniformly sample one each time, since we empirically found that using multiple surrogate models can make attacks stronger.

\textbf{Experimental verification of increased diversity:} 
We quantitatively evaluate that ODS can lead to high diversity in the output space of the target model, as shown in the right two plots of Figure~\ref{figure1}. We use pre-trained Resnet50~\citep{resnet16} and VGG19~\citep{VGG19} models on ImageNet as the target and surrogate models respectively. We calculate and compare the mean pairwise distances of samples with ODS and random Gaussian sampling. The pairwise distance (i.e. diversity) for ODS is 0.79, which is 10 times larger than Gaussian sampling (0.07). Details are in Appendix~\ref{appendix:ap_black_diversity}. We additionally observe that ODS does not produce diversified samples when we use random networks as surrogate models. This indicates that good surrogate models are crucial for transferring diversity.

\begin{algorithm}[tb]
   \caption{Simple Black-box Attack~\citep{Guo19} with ODS }
   \label{alg:black}
\begin{algorithmic}[1]
   \STATE {\bfseries Input:} A targeted image $\bmx$, loss function $L$, a target classifier $\bmf$, a set of surrogate models $\mathcal{G}$
   \STATE {\bfseries Output:} attack result $\bmx_{\text{adv}}$
   \STATE Set the starting point $\bmx_{\text{adv}} = \bmx$ 
   \WHILE {$\bmx_{\text{adv}}$ is not adversary}
   \STATE  Choose a surrogate model $\bmg$ from $\mathcal{G}$, and sample $\bmw_{\text{d}} \sim U(-1,1)^C$
   \STATE  Set $\bmq = \bmv_{\text{ODS}}(\bmx_{\text{adv}},\bmg,\bmw_{\text{d}})$
   \FOR{$\alpha \in \{\epsilon, -\epsilon\}$}
   \IF{$L(\bmx_{\text{adv}}+\alpha \cdot \bmq) > L(\bmx_{\text{adv}})$}
   \STATE   Set $\bmx_{\text{adv}}=\bmx_{\text{adv}}+\alpha \cdot \bmq$ and {\bf break}
   \ENDIF
   \ENDFOR
   \ENDWHILE
\end{algorithmic}
\end{algorithm}

\section{Experiments in white-box settings}
\label{sec_white}
In this section, we show that the diversity offered by ODI can improve white-box attacks for both $\ell_\infty$ and $\ell_2$ distances. 
Moreover, we demonstrate that a simple combination of PGD and ODI achieves new state-of-the-art attack success rates. All experiments are for untargeted attacks.

\subsection{Efficacy of ODI for white-box attacks}
\label{sec_white_various}
We combine ODI with two popular attacks: PGD attack~\citep{madry17} with the $\ell_\infty$ norm and C\&W attack~\citep{cw17} with the $\ell_2$ norm. 
We run these attacks on MNIST, CIFAR-10 and ImageNet. 

\paragraph{Setup}  
We perform attacks against three adversarially trained models from MadryLab\footnote{\url{https://github.com/MadryLab/mnist_challenge} and \url{https://github.com/MadryLab/cifar10_challenge}. We use their secret model.}~\citep{madry17} for MNIST and CIFAR-10 and the Feature Denoising ResNet152 network\footnote{\url{https://github.com/facebookresearch/ImageNet-Adversarial-Training}.}~\citep{featureDenoise19} for ImageNet. For PGD attacks, we evaluate the model accuracy with 20 restarts, where starting points are uniformly sampled over an $\epsilon$-ball for the na\"{i}ve resampling. For C\&W attacks, we calculate the minimum $\ell_2$ perturbation that yields a valid adversarial example among 10 restarts for each image, and measure the average of the minimum perturbations. Note that the original paper of C\&W~\citep{cw17} attacks did not apply random restarts. Here for the na\"{i}ve initialization of C\&W attacks we sample starting points from a Gaussian distribution and clip it into an $\epsilon$-ball (details in Appendix~\ref{appendix:ap_parameter_whiteall}). 

For fair comparison, we test different attack methods with the same amount of computation. Specifically, we compare $k$-step PGD with na\"{i}ve initialization (denoted as PGD-$k$) against ($k$-2)-step PGD with 2-step ODI (denoted as ODI-PGD-($k$-2)). We do not adjust the number of steps for C\&W attacks because the computation time of 2-step ODI are negligible for C\&W attacks. %

\begin{table*}[htb]
\caption{Comparing different white-box attacks. We report model accuracy (lower is better) for PGD and average of the minimum $\ell_2$ perturbations (lower is better) for C\&W. All results are the average of three trials.
}
\begin{center}
\begin{tabular}{c|cc|cc}
\toprule
 & \multicolumn{2}{|c}{PGD } &
\multicolumn{2}{|c}{C\&W} \\ 
model &  na\"{i}ve (PGD-$k$) & ODI (ODI-PGD-($k$-2)) & na\"{i}ve & ODI 
\\ \midrule
MNIST & $90.31\pm 0.02\%$ & $\textbf{90.21}\pm 0.05\%$ &  ${2.27}\pm0.00$ &  $\textbf{2.25}\pm0.01$  \\ 
CIFAR-10 &  $46.06\pm 0.02\%$  &$\textbf{44.45}\pm 0.02\%$ & $0.71\pm0.00$ & $\textbf{0.67}\pm0.00$ \\ 
ImageNet  &  $43.5\pm 0.0\%$  & $\textbf{42.3}\pm 0.0\%$& $1.58\pm0.00$ & $\textbf{1.32}\pm0.01$ \\ 
\bottomrule         
\end{tabular}
\end{center}
\label{tab_Linf}
\end{table*}

\paragraph{Results} 
We summarize all quantitative results in Table~\ref{tab_Linf}. %
Attack performances with ODI are better than na\"{i}ve initialization for all models and attacks. 
The improvement by ODI on the CIFAR-10 and ImageNet models is more significant than on the MNIST model. We hypothesize that this is due to the difference in model non-linearity.  When a target model includes more non-linear transformations,
the difference in diversity between the input and output space could be larger, in which case ODI will be more effective in providing a diverse set of restarts.

\subsection{Comparison between PGD attack with ODI and state-of-the-art attacks}
\label{sec_sota}
To further demonstrate the power of ODI, we perform ODI-PGD against MadryLab's robust models~\citep{madry17} on MNIST and CIFAR-10  
and compare ODI-PGD with state-of-the-art attacks.

\paragraph{Setup}
One state-of-the-art attack we compare with is the well-tuned PGD attack~\citep{MT19}, which achieved 88.21\% accuracy for the robust MNIST model. The other attack we focus on is the MultiTargeted attack~\citep{MT19}, which obtained 44.03\% accuracy against the robust CIFAR-10 model. 
We use all test images on each dataset and perform ODI-PGD under two different settings. One is the same as Section~\ref{sec_white_various}. 
The other is ODI-PGD with tuned hyperparameters, e.g. increasing the number of steps and restarts. Please see Appendix~\ref{appendix:ap_parameter_ODI} for more details of tuning.

\begin{table*}[htbp]
\caption{Comparison of ODI-PGD with state-of-the-art attacks against pre-trained defense models. The complexity rows display products of the number of steps and restarts. Results for ODI-PGD are the average of three trials.
For ODI-PGD, the number of steps is the sum of ODS and PGD steps. 
}
\begin{center}
\setlength{\tabcolsep}{4.5pt}
\begin{tabular}{cc|cccc}
\toprule
model&  & \begin{tabular}{c}ODI-PGD \\ 
(in Sec.~\ref{sec_white_various}) \end{tabular} & \begin{tabular}{c}tuned  \\ODI-PGD  \end{tabular}
  & \begin{tabular}{c}tuned PGD \\~\citep{MT19} \end{tabular}
  & \begin{tabular}{c}MultiTargeted \\~\citep{MT19} \end{tabular} \\ \midrule
 \multirow{2}{*}{MNIST} &accuracy & $90.21\pm 0.05\%$  & $\textbf{88.13}\pm 0.01\%$  & 88.21\% & 88.36\%  \\
 &complexity & 
$40 \times 20$ &  $1000\times 1000 $ & $1000 \times 1800$  &
$1000 \times 1800$    \\ \hline 
\multirow{2}{*}{CIFAR-10}&accuracy&  $44.45\pm 0.02\%$ & $\textbf{44.00} \pm 0.01\%$  & 44.51\%  & 44.03\% \\
  &complexity & $20 \times 20$ & $150 \times 20$ & $1000 \times 180$ & $1000 \times 180$    \\ 
\bottomrule
\end{tabular}
\end{center}
\label{tab_sota}
\end{table*}

\paragraph{Results}
We summarize the comparison between ODI-PGD and state-of-the-art attacks in Table~\ref{tab_sota}. 
Our tuned ODI-PGD reduces the accuracy 
to 88.13\% for the MNIST model, and to 44.00\% for the CIFAR-10 model. These results outperform existing state-of-the-art attacks.

To compare their running time, we report the total number of steps (the number of steps multiplied by the number of restarts) as a metric of complexity, because the total number of steps is equal to the number of gradient computations (the computation time per gradient evaluation is comparable for all gradient-based attacks).
In Table~\ref{tab_sota}, the computational cost of tuned ODI-PGD is smaller than that of state-of-the-art attacks, and especially 50 times smaller on  CIFAR-10. 
Surprisingly, even without tuning ODI-PGD (in the first column) can still outperform  tuned PGD~\citep{MT19}
while also being 
drastically more efficient computationally.

\section{Experiments in black-box settings}
\label{sec_black_experiment}
In this section, we demonstrate that 
black-box attacks combined with ODS significantly reduce the number of queries needed to generate adversarial examples. 
In experiments below, we randomly sample 300 correctly classified images from the ImageNet validation set. We evaluate both untargeted and targeted attacks. For targeted attacks, we uniformly sample target labels.

\subsection{Query-efficiency of score-based attacks with ODS}
\label{sec_black_score}
\subsubsection{Applying ODS to score-based attacks}
\label{sec_black_score_simba}
To show the efficiency of ODS, we combine ODS with the score-based Simple Black-Box Attack (SimBA)~\citep{Guo19}. 
SimBA randomly samples a vector and 
either adds or subtracts the vector to the target image to explore update directions. 
The vector is sampled from a pre-defined set of  orthonormal vectors in the input space. 
These are the discrete cosine transform (DCT) basis vectors in the original paper~\citep{Guo19}. We replace the DCT basis vectors with ODS sampling  (called SimBA-ODS), %

\paragraph{Setup}
We use pre-trained ResNet50 model as the target model 
and select four pre-trained models 
(VGG19, ResNet34, DenseNet121~\citep{densenet}, MobileNetV2~\citep{mobilenetv2}) as surrogate models.
We set the same hyperparameters for SimBA as~\citep{Guo19}: the step size is $0.2$ and the number of iterations (max queries) is 10000 (20000) for untargeted attacks and 30000 (60000) for targeted attacks. As the loss function in SimBA, we employ the margin loss for untargeted attacks and the cross-entropy loss for targeted attacks.

\paragraph{Results}
First, we compare  SimBA-DCT~\citep{Guo19} and SimBA-ODS. Table~\ref{tab_black_score} reports the number of queries and the median $\ell_2$ perturbations. Remarkably, SimBA-ODS reduces the average number of queries by a factor between 2 and 3 compared to SimBA-DCT for both untargeted and targeted settings. This confirms that ODS not only helps white-box attacks, but also leads to significant improvements of query-efficiency in black-box settings. In addition, SimBA-ODS decreases the average perturbation sizes by around a factor of two, which means that ODS helps find better adversarial examples that are closer to the original image. 

\begin{table}[htb]
\caption{Number of queries and size of $\ell_2$ perturbations for score-based attacks.}
\begin{center}
\setlength{\tabcolsep}{3.9pt}
\begin{tabular}{cc|ccc|ccc}
\toprule
&& \multicolumn{3}{c}{untargeted} &\multicolumn{3}{|c}{targeted}  \\ \cline{3-8}
& num. of & success  & average & median $\ell_2$ & success  & average & median $\ell_2$  \\ \
attack& surrogates &  rate & queries &  perturbation &  rate & queries &  perturbation  \\
\midrule
SimBA-DCT~\citep{Guo19} &0& {\bf 100.0\%} & 909 & 2.95 & 97.0\% & 7114 &7.00 \\
SimBA-ODS &4& {\bf 100.0\%} & {\bf 242} & {\bf 1.40} & {\bf 98.3\%} & {\bf 3503} & {\bf 3.55} \\
\bottomrule
\end{tabular}
\end{center}
\label{tab_black_score}
\end{table}

\subsubsection{Comparing ODS with other methods using surrogate models}
\label{sec_black_score_RGF}
We consider another black-box attack that relies on surrogate models: P-RGF~\citep{Cheng19prior}, which improves over the original RGF (random gradient-free) method for gradient estimation. 
P-RGF exploits prior knowledge from surrogate models to estimate the gradient more efficiently than RGF. Since RGF uses random sampling to estimate the gradient, we propose to apply ODS to RGF (new attack named ODS-RGF) and compare it with P-RGF under $\ell_2$ and $\ell_\infty$ norms.

\begin{table}[htbp]
\caption{ Comparison of ODS-RGF and P-RGF on ImageNet. Hyperparameters for RGF are same as~\citep{Cheng19prior} :max queries are 10000, sample size is 10, step size is 0.5 ($\ell_2$) and 0.005 ($\ell_\infty$), and epsilon is $\sqrt{0.001 \cdot 224^2 \cdot 3}$ ($\ell_2$) and 0.05 ($\ell_\infty$). }
\begin{center}
\setlength{\tabcolsep}{3.5pt}
\begin{tabular}{ccc|ccc|ccc}
\toprule
& & &\multicolumn{3}{c}{untargeted} &\multicolumn{3}{|c}{targeted}  \\ \cline{4-9}
  &  &num. of & success& average & median $\ell_2$ & success  &  average & median $\ell_2$   \\ 
norm& attack &surrogates & rate  & queries & perturbation & rate  &  queries & perturbation  \\ 
\midrule
\multirow{3}{*}{$\ell_2$}&RGF & 0& \textbf{100.0\%} & 633 & 3.07 &\textbf{99.3\%} &3141 & 8.23\\
&P-RGF~[25] &1& \textbf{100.0\%} &211 & 2.08 
&{97.0\%} &2296 & 7.03\\
&ODS-RGF& 1& \textbf{100.0\%} &\textbf{133}& \textbf{1.50}
&\textbf{99.3\%} &\textbf{1043} & \textbf{4.47}\\ \hline
\multirow{3}{*}{$\ell_\infty$} &RGF  &0& {97.0\%} & 520 & - 
&{25.0\%} & 2971   & -  \\
&P-RGF~[25]  &1& {99.7\%} &88& - & {65.3\%} & 2123 & -\\
&ODS-RGF& 1& \textbf{100.0\%} &\textbf{74}& - & \textbf{92.0\%} & \textbf{985} & -  \\
\bottomrule
\end{tabular}
\end{center}
\label{fig_black_RGF}
\end{table}

For fair comparison, we use a single surrogate model as in~\citep{Cheng19prior}.
We choose pre-trained ResNet50 model as the target model and ResNet34 model as the surrogate model. 
We give query-efficiency results of both methods in Table~\ref{fig_black_RGF}. The average number of queries required by ODS-RGF is less than that of P-RGF in all settings. This suggests ODS-RGF can estimate the gradient more precisely than P-RGF by exploiting diversity obtained via ODS and  surrogate models. 
The differences between ODS-RGF and P-RGF are significant in targeted settings, since ODS-RGF achieves smaller perturbations than P-RGF (see median perturbation column). 
To verify the robustness of our results, we also ran experiments using VGG19 as a surrogate model and obtained similar results. %

We additionally consider TREMBA~\citep{Huang20TREMBA}, a black-box attack (restricted to the $\ell_\infty$-norm) that is state-of-the-art among those using surrogate models. In TREMBA, a
low-dimensional embedding is learned via surrogate models so as to obtain initial adversarial examples
which
are then updated using a score-based attack. %
Our results show that ODS-RGF combined with SI-NI-DIM~\citep{NesterovTransfer2020}, which is a state-of-the-art transfer-based attack, is comparable to TREMBA even though ODS-RGF is not restricted to the $\ell_\infty$-norm. 
Results and more details are provided in Appendix~\ref{appendix:ap_black_TREMBA}.

\subsubsection{Comparison of ODS with state-of-the-art score-based attacks}
To show the advantage of ODS and surrogate models, 
 we compare SimBA-ODS and ODS-RGF with the Square Attack~\citep{ACFH2019square}, which is a state-of-the-art attack for both $\ell_{\infty}$ and $\ell_2$ norms when surrogate models are not allowed.
For comparison, we regard SimBA as $\ell_2$ bounded attacks: the attack is successful when adversarial $\ell_2$ perturbation is less than a given bound $\epsilon$.
We set $\epsilon=5 \ (\ell_2)$ and $0.05 \ (\ell_\infty)$ as well as other hyperparameters according to the original paper~\citep{ACFH2019square}, except that we set the max number of queries to $20000$ for untargeted attacks and $60000$ for targeted attacks. For ODS-RGF, we use four surrogate models as discussed in Section~\ref{sec_black_score_simba} for SimBA-ODS. %

\begin{table}[htbp]
    \caption{Number of queries for attacks with ODS versus the Square Attack. }
    \begin{center}
    \label{tab_black_square}
    \begin{tabular}{ccc|cc|cc}
    \toprule
    &&& \multicolumn{2}{c}{untargeted} &\multicolumn{2}{|c}{targeted} \\ \cline{4-7}
    && num. of & success   & average &  success   & average \\ 
    norm &attack& surrogates&  rate  & queries &  rate  & queries  \\ \midrule
    \multirow{3}{*}{$\ell_2$}&Square~\citep{ACFH2019square} &0&  {99.7\%} & 647
    & {96.7\%}  & {11647} \\
    &SimBA-ODS &4& {99.7\%} &  {237}  & {90.3\%}  & {2843} \\
    &ODS-RGF & 4& {\bf 100.0\%} & {\bf 144}  & {\bf 99.0\%} & {\bf 1285}\\
     \hline
    \multirow{2}{*}{$\ell_\infty$} &Square~\citep{ACFH2019square} &0&  {\bf 100.0 \%} & {\bf  60} & {\bf 100.0\%}  & {2317} \\
    &ODS-RGF &4&  {\bf 100.0 \%} & 78 & {97.7\%}& {\bf 1242} \\
    \bottomrule
    \end{tabular}
    \end{center}
\end{table}

As shown in Table~\ref{tab_black_square}, 
the number of queries required for ODS-RGF and SimBA-ODS are lower than that of the Square Attack under the $\ell_2$ norm. The improvement is especially large for ODS-RGF. The difference between ODS-RGF and SimBA-ODS mainly comes from different base attacks (i.e., RGF and SimBA).
For the $\ell_\infty$ norm setting, ODS-RGF is comparable to the Square Attack. We hypothesize that the benefit of estimated gradients by RGF decreases under the $\ell_\infty$ norm due to the sign function. However, because ODS can be freely combined with many base attacks, a stronger base attack is likely to  further improve query-efficiency.

\subsection{Query-efficiency of decision-based attacks with ODS}
\label{sec_black_decision}
We demonstrate that ODS also improves query-efficiency for decision-based attacks. We combine ODS with the decision-based Boundary Attack~\citep{Brendel18}. 
The Boundary Attack starts from an image which is adversarial, and iteratively updates the image to find smaller perturbations. 
To generate the update direction, the authors of~\citep{Brendel18} sampled a random noise vector from a Gaussian distribution $\mathcal{N}(\mathbf{0}, \mathbf{I})$ each step.
We replace this random sampling procedure with sampling by ODS (we call the new method Boundary-ODS). We give the pseudo-code of Boundary-ODS in Algorithm~\ref{alg:ap_boundary} (in the Appendix).

\paragraph{Setup}
We use the same settings as the previous section for score-based attacks: 300 validation images on ImageNet, pre-trained ResNet50 target model, and four pre-trained surrogate models.
We test on both untargeted and targeted attacks. In targeted settings, we give randomly sampled images with target labels as initial images.
We use the implementation in Foolbox~\citep{foolbox2017} for Boundary Attack with default parameters, which is more efficient than the original implementation.

We also compare Boundary-ODS with two state-of-the-art decision-based attacks: 
the HopSkipJump attack~\citep{chen2019hop} and the Sign-OPT attack~\citep{cheng20sign}. We use the implementation in ART~\citep{art2018} for HopSkipJump and the author's implementation for Sign-OPT. We set default hyperparameters for both attacks.

\begin{figure}[htbp]
\centering

    \begin{tabular}{cc}
      \begin{minipage}{0.35\hsize} %
        \begin{center}
          \includegraphics[width=1.0\textwidth]{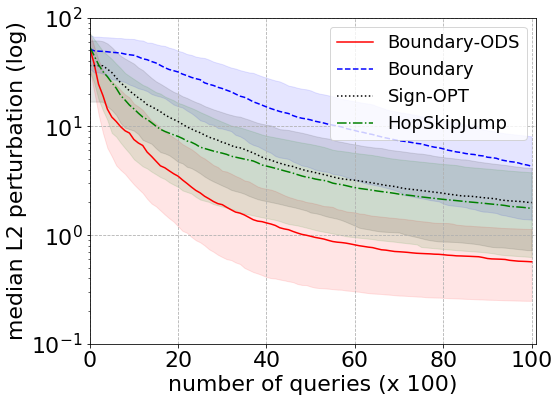}
        \end{center}
      \end{minipage}
        &
      \begin{minipage}{0.35\hsize}
        \begin{center}
          \includegraphics[width=1.0\textwidth]{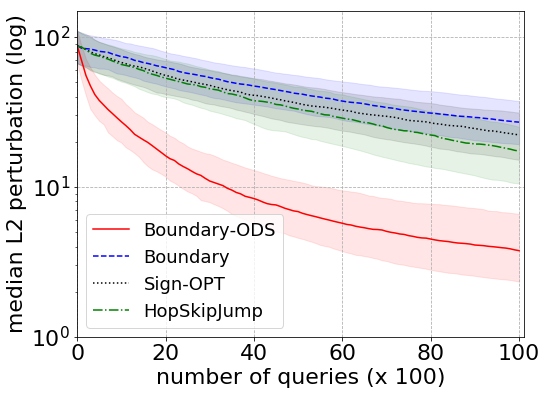}
        \end{center}
      \end{minipage}
      \\
      Untargeted & Targeted 
    \end{tabular}
    \caption{Relationship between median $\ell_2$ perturbations and the number of queries for decision-based attacks. Error bars show 25th and 75th percentile of $\ell_2$ perturbations. } %
    \label{fig_black_decision}
\end{figure}

\paragraph{Results}
Table~\ref{tab_black_decision} summarizes the median sizes of $\ell_2$ adversarial perturbations obtained with a fixed number of queries. Clearly, Boundary-ODS significantly improves query-efficiency compared to the original Boundary Attack. In fact, Boundary-ODS outperforms state-of-the-art attacks: it decreases the median $\ell_2$ perturbation at 10000 queries to less than one-third of previous best untargeted attacks and less than one-fourth of previous best targeted attacks. 
We additionally describe the relationship between median $\ell_2$ perturbations and the number of queries in Figure~\ref{fig_black_decision}. Note that Boundary-ODS outperforms other attacks, especially in targeted settings. Moreover, Boundary-ODS only needs fewer than 3500 queries to achieve the adversarial perturbation obtained by other attacks with 10000 queries.

\begin{table}[htb]
\caption{Median $\ell_2$ perturbations for Boundary-ODS and decision-based state-of-the-art attacks.}
\begin{center}
\begin{tabular}{cc|ccc|ccc}
\toprule
& & \multicolumn{6}{c}{number of queries} \\
& num. of & \multicolumn{3}{c}{untargeted} &\multicolumn{3}{|c}{targeted} \\ 
attack &surrogates& 1000 & 5000 & 10000 & 1000 & 5000 & 10000 \\ \midrule
Boundary~\citep{Brendel18} & 0& 45.07& 11.46 & 4.30 & 73.94 &41.88 &27.05  \\
Boundary-ODS & 4&  {\bf 7.57} & {\bf 0.98} & {\bf 0.57} & {\bf 27.24} & {\bf 6.84} & {\bf 3.76}\\
HopSkipJump~\citep{chen2019hop} & 0& 14.86 &3.50 & 1.79 & 65.88 & 33.98 & 18.25 \\
Sign-OPT~\citep{cheng20sign} & 0& 21.73 & 3.98 & 2.01 & 68.75 & 36.93&22.43\\ 
\bottomrule
\end{tabular}
\end{center}
\label{tab_black_decision}
\end{table}

\subsection{Effectiveness of ODS with out-of-distribution images}
\label{sec_black_limited}
Although several studies use prior knowledge from surrogate models to improve performance of black-box attacks, there is a drawback---those approaches require a dataset to train surrogate models. 
In reality, it is typically impossible to obtain the same dataset used for training the target model. We show that ODS is applicable even when we only have a limited dataset that is out-of-distribution (OOD) and may contain only images with irrelevant labels.

We select 100 ImageNet classes which do not overlap with the classes used in the experiments of Section~\ref{sec_black_decision}.  
We train surrogate models using an OOD training dataset with these 100 classes. 
We train five surrogate models with the same ResNet18 architecture because multiple surrogate models provide diversified directions. 
Then, we run Boundary-ODS with the trained surrogate models under the same setting as Section~\ref{sec_black_decision}. As shown in Table~\ref{tab_black_limited1}, although Boundary-ODS with the OOD training dataset underperforms Boundary-ODS with the full dataset, it is still significantly better than the original Boundary Attack with random sampling. This demonstrates that the improved diversity achieved by ODS improves black-box attacks even if we only have OOD images to train a surrogate.

\begin{table}[htb]
\caption{Median $\ell_2$ perturbations for Boundary-ODS with surrogate models trained on OOD images.}
\begin{center}
\begin{tabular}{c|ccc|ccc}
\toprule
&  \multicolumn{6}{c}{number of queries} \\
&  \multicolumn{3}{c}{untargeted} &\multicolumn{3}{|c}{targeted} \\ 
attack & 1000 & 5000 & 10000 & 1000 & 5000 & 10000 \\ \midrule
Boundary~\citep{Brendel18} & 45.07& 11.46 & 4.30 & 73.94 &41.88 &27.05  \\
Boundary-ODS (OOD dataset) &  {\bf 11.27}& {\bf 1.63}& {\bf 0.98}&{\bf 41.67}&{\bf 13.72} & {\bf 8.39}\\ \hline
Boundary-ODS (full dataset in Sec.~\ref{sec_black_decision}) &  7.57& 0.98& 0.57& 27.24& 6.84& 3.76\\
\bottomrule
\end{tabular}
\end{center}
\label{tab_black_limited1}
\end{table}

\section{Related works}
\label{sec_related}

The closest approach to ours is the white-box MultiTargeted attack~\citep{MT19}. This attack changes the target class of attacks per restart, and it can be regarded as a method which aims to obtain more diversified starting points. However, MultiTargeted attack is limited to the setting of $\ell_p$-bounded white-box attacks. In contrast, ODS can be applied to more general white- and black-box attacks. In addition, ODS does not require the original class of the target image, therefore it is more broadly applicable. Further discussion are in Appendix~\ref{appendix:ap_multitargeted}.
Another related work is Interval Attack~\citep{wang19symbolic} which generates diverse starting points by leveraging symbolic interval propagation. While Interval Attack shows good performances against MNIST models, the attack is not scalable to large models. 

ODS utilizes surrogate models, which are commonly used for black-box attacks. Most previous methods exploit surrogate models to estimate gradients of the loss function on the target model~\citep{trans_papernot17,trans_liu17,Brunner19GuessingSmart,Cheng19prior,subspaceattack,Cai19transferSMBdirect}.
Some recent works exploit surrogate models to train other models~\citep{Du2020Query-efficient,Huang20TREMBA}  
or update surrogate models during attacks~\citep{Suya20hybridBlack}. Integrating ODS with these training-based methods is an interesting direction for future work.

\section{Conclusion}
We propose ODS, a new sampling strategy for white- and black-box attacks. 
By generating more diverse perturbations as measured in the output space, ODS can create more effective starting points for white-box attacks. Leveraging surrogate models, ODS also improves the exploration of the output space for black-box attacks. Moreover, ODS for black-box attacks is applicable even if the surrogate models are trained with out-of-distribution datasets. Therefore, black-box attacks with ODS are more practical than other black-box attacks using ordinary surrogate models. Our empirical results demonstrate that ODS with existing attack methods outperforms state-of-the-art attacks in various white-box and black-box settings. 

While we only focus on ODS with surrogate models trained with labeled datasets, ODS may also work well using unlabeled datasets, which we leave as future work. One additional direction is to improve the efficiency of ODS by selecting suitable surrogate models with reinforcement learning. 

\section*{Broader Impact}
The existence of adversarial examples is a major source of concern for machine learning applications in the real world. For example, imperceptible perturbations crafted by malicious attackers could deceive safety critical systems such as autonomous driving and facial recognition systems. Since adversarial examples exist not only for images, but also for other domains such as text and audio, the 
potential impact is large. 
Our research provides new state-of-the-art black-box adversarial attacks in terms of query-efficiency and makes adversarial attacks more practical and strong. 
While all experiments in this paper are for images, the proposed method is also applicable to other modalities. Because of this, our research could be used in harmful ways by malicious users. 

On the positive side, strong attacks are necessary to develop robust machine learning models. 
For the last few years, several researchers have proposed adversarial attacks which break previous defense models.
In response to these strong attacks, new and better defense mechanisms have been developed. 
It is this feedback loop between attacks and defenses that advances the field. %
Our research not only provides a state-of-the-art attack, 
but also sheds light on a new perspective, namely the importance of diversity, for improving adversarial attacks. This may have a long term impact on inspiring more effective defense methods.

\section*{Acknowledgements and Disclosure of Funding}
This research was supported in part by AFOSR (FA9550-19-1-0024), NSF (\#1651565, \#1522054, \#1733686), ONR, and FLI.

\bibliography{arxiv_camera}
\bibliographystyle{unsrt}

\newpage
\renewcommand{\thetable}{\Alph{table}}
\renewcommand{\thefigure}{\Alph{figure}}
\renewcommand{\thealgorithm}{\Alph{algorithm}}
\renewcommand{\theequation}{\Alph{equation}}
\setcounter{table}{0}
\setcounter{figure}{0}
\setcounter{equation}{0}
\setcounter{algorithm}{0}
\appendix

\section{Pseudo-code of proposed methods}
\label{appendix:ap_pseudocode}
In this section, we provide the pseudo-codes of methods proposed in the main paper. First, Algorithm~\ref{alg:ap_white} shows the pseudo-code of ODI for white-box attacks in Section~\ref{sec_ODS_white}. The line 4-6 in the algorithm describes the iterative update by ODI.

\begin{algorithm}[htbp]
   \caption{Initialization by ODS (ODI) for white-box attacks }
   \label{alg:ap_white}
\begin{algorithmic}[1]
   \STATE {\bfseries Input:} A targeted image $\bmx_{org}$, a target classifier $\bmf$, perturbation set $B(\bmx_{org})$, number of ODI steps $N_{\text{ODI}}$, step size $ \eta_{\text{ODI}}$, number of restarts $N_R$
   \STATE {\bfseries Output:} Starting points $\{x^{start}_i \}$ for adversarial attacks
   \FOR{$i=1$ {\bfseries to} $N_R$}
   \STATE Sample $\bmx_{0}$ from $B(\bmx_{org})$, and sample $\bmw_{\text{d}} \sim U(-1,1)^C$
   \FOR{$k=0$ {\bfseries to} $N_{\text{ODI}}-1$}
   \STATE  $\bmx_{k+1}  \gets \mathrm{Proj}_{B(\bmx_{org})} \left( x_{k} + \eta_{\text{ODI}} \, \mathrm{sign}(\bmv_{\text{ODS}}(\bmx_k,\bmf,\bmw_{\text{d}}) ) \right)$
   \ENDFOR
   \STATE $\bmx^{start}_i \gets x_{N_{\text{ODI}}}$
   \ENDFOR
\end{algorithmic}
\end{algorithm}

We also describe the algorithm of Boundary-ODS, used in Section~\ref{sec_black_decision} of the main paper.  Algorithm~\ref{alg:ap_boundary} shows pseudo-code of Boundary-ODS. 
The original Boundary Attack~\citep{Brendel18} first sampled a random noise vector $\bmq$ from a Gaussian distribution $\mathcal{N}(\mathbf{0}, \mathbf{I})$ and then orthogonalized the vector to keep the distance from the original image (line 7 in Algorithm~\ref{alg:ap_boundary}).
After that, the attack refined the vector $\bmq$ to reduce the distance from the original image such that the following equation holds:
\begin{equation}
\label{eq_decision_update}
d(\bmx,\bmx_{adv}) - d(\bmx,\bmx_{adv}+\bmq) = \epsilon \cdot d(\bmx,\bmx_{adv})
\end{equation}
where $d(a,b)$ is the distance between $a$ and $b$.
We replace the random Gaussian sampling to ODS as in the line 5 and 6 of Algorithm~\ref{alg:ap_boundary}.
Sampled vectors by ODS yield large changes for outputs on the target model and increase the probability that the updated image is adversarial (i.e. the image satisfies the line 9 of Algorithm~\ref{alg:ap_boundary}), so ODS makes the attack efficient.

\begin{algorithm}[htbp]
   \caption{Boundary Attack~\citep{Brendel18} with sampling update direction by ODS }
   \label{alg:ap_boundary}
\begin{algorithmic}[1]
   \STATE {\bfseries Input:} A targeted image $\bmx$, a label $y$, a target classifier $\bmf$, a set of surrogate models $\mathcal{G}$
   \STATE {\bfseries Output:} attack result $\bmx_{adv}$
   \STATE Set the starting point $\bmx_{adv} = \bmx$ which is adversary
   \WHILE {$k<$ number of steps}
   \STATE  Choose a surrogate model $\bmg$ from $\mathcal{G}$, sample $\bmw_{\text{d}} \sim U(-1,1)^C$
   \STATE  Set $\bmq = \bmv_{\text{ODS}}(\bmx_{adv},\bmg,\bmw_{\text{d}})$
   \STATE  Project $\bmq$ onto a sphere around the original image $\bmx$
   \STATE Update $\bmq$ with a small movement toward the original image $\bmx$ such that Equation~(\ref{eq_decision_update}) holds
   \IF{$\bmx_{adv}+\bmq$ is adversarial}
   \STATE   Set $\bmx_{adv}=\bmx_{adv}+\bmq$ 
   \ENDIF
   \ENDWHILE
\end{algorithmic}
\end{algorithm}

\section{Details of experiment settings}
\label{appendix:ap_parameter}

\subsection{Hyperparameters and settings for attacks in Section~\ref{sec_white_various}}
\label{appendix:ap_parameter_whiteall}
We describe hyperparameters and settings for PGD and C\&W attacks in Section~\ref{sec_white_various}. 

Multiple loss functions $L(\cdot)$ can be used for PGD attacks, including the cross-entropy loss, and the margin loss defined as $\max_{i \neq y} f_{i}(\bmx) - f_{y}(\bmx)$. %
We use the margin loss for PGD attacks to make considered attacking methods stronger.

PGD attacks have three hyperparameters: pertubation size $\epsilon$, step size $\eta$ and number of steps $N$. 
We chose $\epsilon=0.3,8/255,4/255$, $\eta= 0.02, 2/255, 0.5/255$ and $N=40,20,50$ for MadryLab (MNIST), MadryLab (CIFAR-10), ResNet152 Denoise (ImageNet), respectively. We use the whole test set except for ImageNet, where the first 1000 test images are used. 

For C\&W attacks, we define na\"{i}ve random initialization to make sure the starting points are within an $\ell_2$ $\epsilon$-radius ball: we first sample Gaussian noise $\bmw \sim  \mathcal{N}(\mathbf{0}, \mathbf{I})$ and then add the clipped noise $\epsilon \cdot \bmw / \|\bmw\|_2$ to an original image. We set the perturbation radius of initialization $\epsilon$ by reference to attack bounds in other studies: $\epsilon= 2.0, 1.0, 5.0$ for MadryLab (MNIST), MadryLab (CIFAR-10), ResNet152 Denoise (ImageNet), respectively.
we also set hyperparameters of C\&W attacks as follows: max iterations are 1000 (MNIST) and 100 (CIFAR-10 and ImageNet), search step is 10, learning rate is 0.1, and initial constant is 0.01. The attack is performed for the first 1000 images (MNIST and CIFAR-10) and the first 500 images (ImageNet).

\subsection{Hyperparameter tuning for tuned ODI-PGD in Section~\ref{sec_sota} }
\label{appendix:ap_parameter_ODI}
We describe hyperparameter tuning for our tuned ODI-PGD in Section~\ref{sec_sota}. 
We summarize the setting in Table~\ref{tab_para}. 

\begin{table*}[ht]
\caption{Hyperparameter setting for tuned ODI-PGD in Section~\ref{sec_sota}.}
\label{tab_para}
\begin{center}
\setlength{\tabcolsep}{4.5pt}
\begin{tabular}{c|cc|ccc}
\toprule
  & \multicolumn{2}{c|}{\text{ODI}}  & \multicolumn{3}{c}{PGD} \\ 
model & \begin{tabular}{c} total step \\ $N_{\text{ODI}}$ \end{tabular}  & \begin{tabular}{c} step size \\ $\eta_{\text{ODI}}$ \end{tabular} & 
  optimizer & 
 \begin{tabular}{c} total step \\ $N$ \end{tabular}   & 
 \begin{tabular}{c} step size (learning rate) \\ $\eta_k$  \end{tabular}                    \\ \midrule
\begin{tabular}{c}MNIST
\end{tabular} & 50 & 
    0.05   & Adam & 950  & 
  $\begin{array}{l@{~}l}
    0.1 & (k < 475) \\
    0.01 & (475 \leq k < 712) \\
    0.001 & (712 \leq k)
  \end{array}$ \\ \hline
\begin{tabular}{c}CIFAR-10
\end{tabular}  & 10  & 
    8/255 & \begin{tabular}{c}sign \\ function \end{tabular} & 140 & 
  $\begin{array}{l@{~}l}
    8/255 & (k < 46) \\
    0.8/255  & (46 \leq k <  92) \\
    0.08/255 & (92 \leq k)
  \end{array}$ \\ 
\bottomrule
\end{tabular}
\end{center}
\end{table*}
For ODI, we increase the number of ODI step $N_{\text{ODI}}$ to obtain more diversified inputs than ODI with $N_{\text{ODI}}=2$. In addition, we make step size $\eta_{\text{ODI}}$ smaller than $\epsilon$ on MNIST, because $\epsilon$-ball with $\epsilon=0.3$ is large and $\eta_{\text{ODI}}=0.3$ is not suitable for seeking the diversity within the large  $\epsilon$-ball. 
In summary, we set $(N_{\text{ODI}},\eta_{\text{ODI}})=(50,0.05), (10, 8/255)$ for the MNIST model and the CIFAR-10 model, respectively.

We tune hyperparameters of PGD based on Gowal et al.~\citep{MT19}.  
While several studies used the sign function to update images for the PGD attack, some studies \citep{SPSA18,MT19} reported that updates by Adam optimizer~\citep{Adam15} brought better results than the sign function. Following the previous studies~\citep{SPSA18,MT19}, we consider the sign function as an optimizer and the choice of an optimizer as a hyperparameter. 
We use Adam for the PGD attack on the MNIST model and the sign function on the CIFAR-10 model. 

We adopt scheduled step size instead of fixed one. 
Because we empirically found that starting from large step size brings better results, we set the initial step size $\eta_0$ as $\eta_0=\epsilon$ on CIFAR-10. We update step size at $k=0.5N, 0.75N$ on MNIST and $k=N/3, 2N/3$ for on CIFAR-10. 
When we use Adam, step size is considered as learning rate.
Finally, we set PGD step $N$ as $N_{\text{ODI}}+N=1000$ on MNIST and $150$ on CIFAR-10.

\subsection{Setting for training on ImageNet in Section~\ref{sec_black_limited}}
\label{appendix:ap_parameter_blackOOD}
We describe the setting of training of surrogate models on ImageNet in the experiment of Section~\ref{sec_black_limited}.
We use the implementation of training provided in PyTorch with default hyperparameters.
Namely, training epochs are 90 and learning rates are changed depending on epoch: 0.1 until 30 epochs, 0.01 until 60 epochs, 0.001 until 90 epochs. Batch size is 256 and weight decay 0.0001.

\section{Additional results and experiments for ODI with white-box attacks}
\label{appendix:ap_white}

\subsection{Diversity offered by ODI}
\label{appendix:ap_white_diversity}
\label{sec_append_white_diversity}
We empirically demonstrate that ODI can find a more diverse set of starting points than random uniform initialization, as pictorially shown in the left figures of Figure~\ref{figure1} of the main paper. 

As an example of target models, we train a robust classification model using adversarial training~\citep{madry17} on CIFAR-10. We adopted popular hyperparameters for adversarial training under the $\ell_\infty$ PGD attack on CIFAR-10: perturbation size $\epsilon = 8/255$, step size $\eta=2/255$, and number of steps $N=10$. Training epochs are 100 and learning rates are changed depending on epoch: 0.1 until 75 epochs, 0.01 until 90 epochs, and 0.001 until 100 epochs. Batch size is 128 and weight decay 0.0002.

On the target model, we quantitatively evaluate the diversity of starting points by each initialization in terms of pairwise distances of output values $\bmf(\bmx)$. Each initialization is bounded within $\ell_\infty$ $\epsilon$-ball with $\epsilon=8/255$.
We pick 100 images on CIFAR-10 and run each initialization 10 times to calculate the mean pairwise distances among outputs for different starting points. As a result, the mean pairwise distance obtained from ODI is 6.41, which is about 15 times larger than that from uniform initialization (0.38). This corroborates our intuition that starting points obtained by ODI are more diverse than uniform initialization. We note that PGD does not generate diverse samples. When we use PGD with 2 steps as an initialization, the mean pairwise distance is only 0.43. 

\begin{figure*}[htbp]
\centering
    \begin{tabular}{ccc}

      \begin{minipage}{0.35\hsize}
        \begin{center}
          \includegraphics[width=1.0\textwidth]{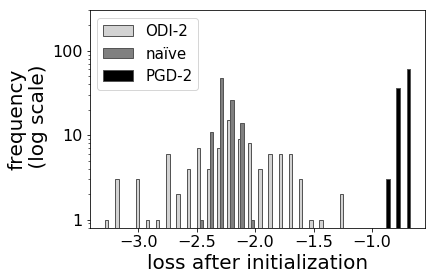}
        \end{center}
      \end{minipage}
      \begin{minipage}{0.35\hsize}
        \begin{center}
          \includegraphics[width=1.0\textwidth]{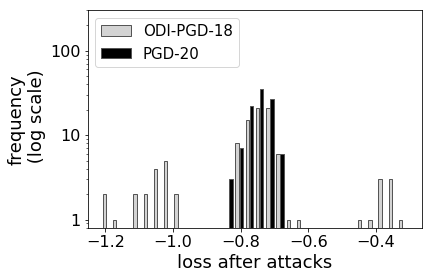}
        \end{center}
      \end{minipage}
      \hspace{0.3cm}
      \begin{minipage}{0.23\hsize}
        \begin{center}
          \includegraphics[width=0.9\textwidth]{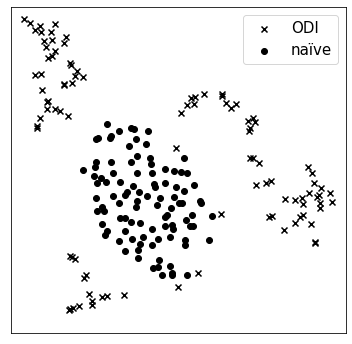}
        \end{center}
      \end{minipage}
    \end{tabular}
    \caption{(Left): Histogram of loss values evaluated at starting points by ODI, na\"{i}ve uniform initialization and PGD. PGD-2 means 2-step PGD with na\"{i}ve initialization. The loss function is the margin loss. (Right) Histogram of loss values after attacks with 20 total steps. ODI-PGD-18 means 18-step PGD with 2-step ODI. 
    (Right): Embedding for starting points sampled on each initialization produced by t-SNE.
    }
    \label{fig_div_visualize}
\end{figure*}

We also visualize the diversity offered by ODI. 
First, we focus on loss histogram of starting points by ODI and na\"{i}ve uniform initialization. We pick an image from the CIFAR-10 test dataset and run each initialization 100 times. Then, we calculate loss values for starting points to visualize their diversity in the output space. The left panel of Figure~\ref{fig_div_visualize} is the histogram of loss values for each initialization. We can easily observe that images from na\"{i}ve initialization concentrate in terms of loss values (around $-1.0$), whereas images from ODI-2 (ODI with 2 steps) are much more diverse in terms of the loss values. We also observe that images from PGD-2 also take similar loss values. 
By starting attacks from these initial inputs, we obtain the histogram of loss values in the center panel of Figure~\ref{fig_div_visualize}. We can observe that ODI-PGD generates more diverse results than PGD with na\"{i}ve initialization (PGD-20).

In addition, we apply 
t-SNE~\citep{tSNE} to the output logits for starting points by each initialization.
We visualize the embedding produced by t-SNE in the right panel of Figure~\ref{fig_div_visualize}.
As expected, starting points produced by ODI are more diversified than those by na\"{i}ve initialization.

\subsection{Analysis of the sensitivity to hyperparameters of ODI}
\label{appendix:ap_white_sensitivity}
For ODI, we mainly set the number of ODI steps $N_{\text{ODI}}=2$ and step size $\eta_{\text{ODI}}=\epsilon$. 
To validate the setting, we confirm that ODI-PGD is not sensitive to these hyperparameters. 
We attack adversarially trained models on CIFAR-10 introduced in Section~\ref{sec_append_white_diversity}, and adopt the same attack setup for ODI-PGD on CIFAR-10 as Section~\ref{sec_white_various}.
We test $N_{\text{ODI}}=2,4,8,16$ and $\eta_{\text{ODI}} = \epsilon, \epsilon/2, \epsilon/4, \epsilon/8$, but 
 exclude patterns with  $N_{\text{ODI}} \cdot \eta_{\text{ODI}} < 2 \epsilon$ to make $N_{\text{ODI}} \cdot \eta_{\text{ODI}}$ larger than or equal to the diameter of the $\epsilon$-ball. 
We calculate the mean accuracy for five repetitions of the attack, each with 20 restarts.

\begin{table}[htb]
\caption{The sensitivity to the number of ODI steps $N_{\text{ODI}}$ and step size $\eta_{\text{ODI}}$.  
We repeat each experiment 5 times to calculate statistics.
}
\begin{center}
\begin{tabular}{cc|ccc}
\toprule
$N_{\text{ODI}}$ & $\eta_{\text{ODI}}$ & mean & max & min  \\ \midrule
2 &$\epsilon$ & 44.46\% &44.50\% & 44.45\% \\
4 &$\epsilon /2$ & 44.47\% &44.50\% & 44.42\% \\
4 &$\epsilon $ & 44.42\% &44.48\% & 44.40\% \\
8 &$\epsilon /4$ & 44.47\% &44.52\% & 44.44\% \\
8 &$\epsilon /2$ & 44.42\% &44.48\% & 44.36\% \\
8 &$\epsilon $ &  44.46\% &44.49\% & 44.42\% \\
16 &$\epsilon /8$ & 44.46\% &44.50\% & 44.43\% \\
16 &$\epsilon /4$ & 44.46\% &44.50\% & 44.40\% \\
16 &$\epsilon /2$ &  44.45\% &44.48\% & 44.43\% \\
16 &$\epsilon$ &  44.44\% &44.47\% & 44.41\% \\
 \bottomrule
\end{tabular}
\end{center}
\label{tab_sensi}
\end{table}

Table~\ref{tab_sensi} shows the mean accuracy under ODI-PGD for different hyperparameters. 
The maximum difference in the mean accuracy among different hyperparameters of ODI is only 0.05\%.
Although large $N_{\text{ODI}}$ and $\eta_{\text{ODI}}$ will be useful to find more diversified starting points, the performance of ODI is not very sensitive to hyperparameters.
Thus, we restrict $N_{\text{ODI}}$ to a small value to give fair comparison in terms of computation time as much as possible. 
Table~\ref{tab_sensi} also shows that the difference between the maximum and minimum accuracy is about 0.1\% for all hyperparameter pairs. This result supports the stability of ODI.

\subsection{Accuracy curve for adversarial attacks with ODI}
\label{appendix:ap_white_accuracycurve}
In Section~\ref{sec_white}, we experimentally represented that the diversity offered by ODI improved white-box $\ell_\infty$ and $\ell_2$ attacks. 
we describe the accuracy curve with the number of restarts for attacks with ODI and na\"{i}ve initialization.

\begin{figure*}[htbp]
\centering
    \begin{tabular}{cccc}
      \rotatebox[origin=c]{90}{PGD}&
      \begin{minipage}{0.28\hsize}
        \begin{center}
          \includegraphics[width=1.0\textwidth]{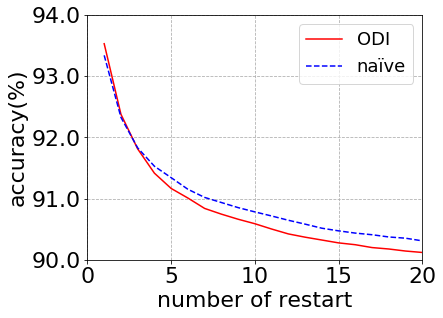}
        \end{center}
      \end{minipage}

      &
      \begin{minipage}{0.28\hsize}
        \begin{center}
          \includegraphics[width=1.0\textwidth]{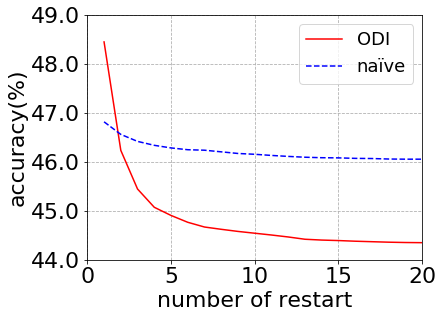}
        \end{center}
      \end{minipage}
      &
      \begin{minipage}{0.28\hsize}
        \begin{center}
          \includegraphics[width=1.0\textwidth]{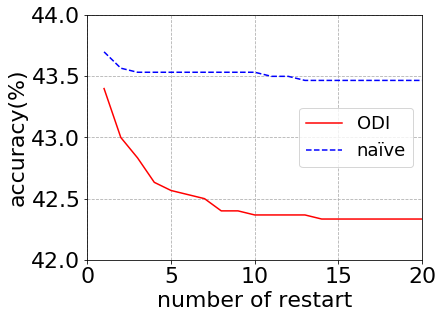}
        \end{center}
      \end{minipage}
      \\
      \rotatebox[origin=c]{90}{C\&W}&
      \begin{minipage}{0.28\hsize}
        \begin{center}
          \includegraphics[width=1.0\textwidth]{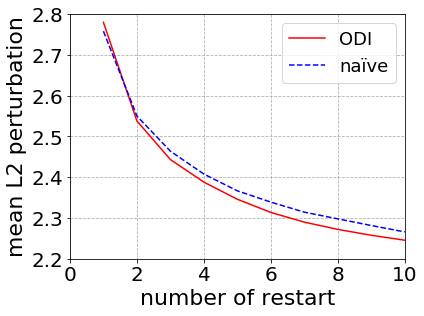}
        \end{center}
      \end{minipage}
      &
      \begin{minipage}{0.28\hsize}
        \begin{center}
          \includegraphics[width=1.0\textwidth]{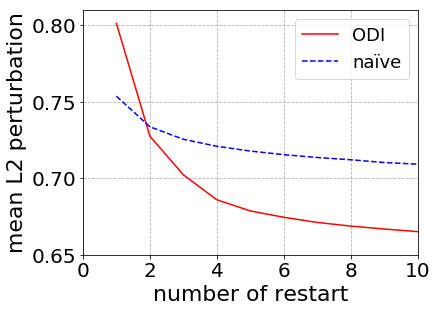}
        \end{center}
      \end{minipage}
      &
      \begin{minipage}{0.28\hsize}
        \begin{center}
          \includegraphics[width=1.0\textwidth]{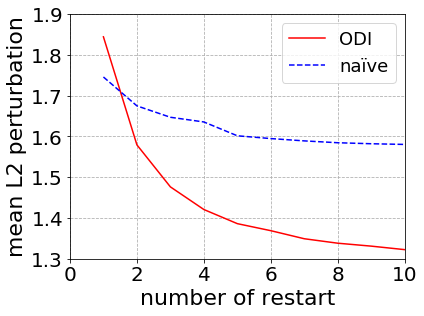}
        \end{center}
      \end{minipage}
      \\ 
        &   \begin{tabular}{c} MNIST 
           \end{tabular}&
           \begin{tabular}{c} CIFAR-10 
           \end{tabular}&
           \begin{tabular}{c} ImageNet 
           \end{tabular}
    \end{tabular}
    \caption{The attack performance against number of restarts for attacks with ODI. (Top): the model accuracy for PGD, (Bottom):  the average of minimum $\ell_2$ perturbations for C\&W. %
    }%
    \label{fig_Linf}
\end{figure*}

Figure~\ref{fig_Linf} shows how the attack performance improves as the number of restarts increases in the experiment of Section~\ref{sec_white_various}. 
Attacks with ODI outperforms those with na\"{i}ve initialization with the increase of restarts in all settings. These curves further corroborate that restarts facilitate the running of attack algorithms, and ODI restarts are more effective than na\"{i}ve ones. We note that the first restart of ODI is sometimes worse than na\"{i}ve initialization. It is because diversity can cause local optima, i.e. random directions of ODI are not always useful. With the increase of restarts, at least one direction is useful and the accuracy drops.

\begin{figure}[ht]
\centering

    \begin{tabular}{cc}
      \begin{minipage}{0.45\hsize}
        \begin{center}
          \includegraphics[width=.9\textwidth]{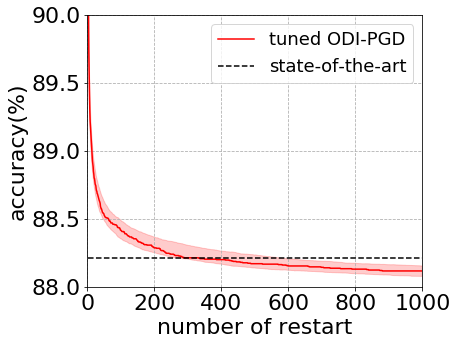}
        \end{center}
      \end{minipage}
        &
      \begin{minipage}{0.45\hsize}
        \begin{center}
          \includegraphics[width=.9\textwidth]{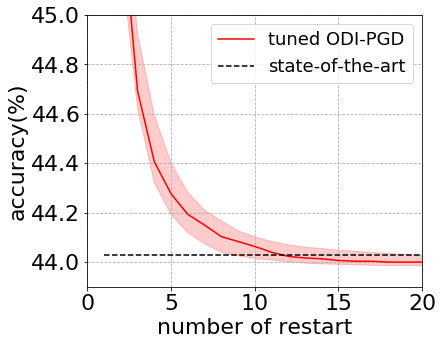}
        \end{center}
      \end{minipage}
      \\
      MadryLab (MNIST) & MadryLab (CIFAR-10) 
    \end{tabular}
    \caption{Model accuracy under tuned ODI-PGD and the current state-of-the-art attacks~\citep{MT19}. 
    The solid lines represent values from Table~\ref{tab_sota} and 
    the error bars show 95\% confidence intervals. }
    \label{fig_SOTAmadry}
\end{figure}

Next, we describe the accuracy curve for the comparison between state-of-the-are attacks and ODI-PGD in Section~\ref{sec_sota}. 
To emphasize the stability of the improvement, 
we evaluate the confidence intervals of our results against MadryLab's MNIST and CIFAR-10 models. We run tuned ODI-PGD attack with 3000 restarts on MNIST and 
100 restarts on CIFAR-10. 
Then, we sample 1000 runs on MNIST and 20 runs on CIFAR-10 from the results to evaluate the model accuracy, 
and re-sample 100 times to calculate statistics. 
Figure~\ref{fig_SOTAmadry} shows the accuracy curve under tuned ODI-PGD. 
We observe that confidence intervals become tighter as the number of restarts grows, and tuned ODI-PGD consistently outperforms the state-of-the-art attack after 1000 restarts on MNIST and 20 restarts on CIFAR-10.

\subsection{Tighter estimation of robustness for various models}
\label{appendix:ap_white_recent}
One important application of powerful adversarial attacks is to evaluate and compare different defense methods. In many previous works on defending against adversarial examples, PGD attack with na\"{i}ve uniform initialization (called na\"{i}ve-PGD) is a prevailing benchmark and its attack success rate is commonly regarded as a tight estimation on (worst-case) model robustness. In this section, we conduct a case study on six recently published defense methods~\citep{UAT19,carmon19,scatter19,metric19,free19,YOPO19} to show that ODI-PGD outperforms na\"{i}ve-PGD in terms of upper bounding the worst model accuracy under all possible attacks.

\paragraph{Setup}
We use pre-trained models from four of those studies,
and train the other two models~\citep{free19,YOPO19}
 using the settings and architectures described in their original papers. We run attacks with $\epsilon = 8/255$ on all test images. Other attack settings are the same as the experiment for CIFAR-10 in Section~\ref{sec_white_various}. Apart from comparing ODI-PGD and na\"{i}ve-PGD, we also evaluate PGD attack without restarts (denoted as $\text{PGD}_{1}$) as it is adopted in several existing studies \citep{UAT19,carmon19,scatter19,YOPO19}.

\begin{table*}[ht]
\caption{Accuracy of models after performing ODI-PGD and na\"{i}ve-PGD attacks against recently proposed defense models.}
\begin{center}
\begin{tabular}{c|ccc|cc}
\toprule
model& (1) $\text{PGD}_{1}$ & (2)  $\text{na\"{i}ve-PGD}$ 
& (3) $\text{ODI-PGD}$ 
&(1)$-$(2) &(2)$-$(3)
\\ \midrule
UAT~\citep{UAT19} & 62.63\% & 61.93\% & {\bf 57.43\%} & 0.70\% & 4.50\% \\
RST~\citep{carmon19} & 61.17\% & 60.77\% &  {\bf 59.93\%} & 0.40\% &  0.84\%  \\
Feature-scatter~\citep{scatter19} & 59.69\% & 56.49\% &  {\bf 39.52\%} & 3.20\% &  16.97\%  \\
Metric learning~\citep{metric19} & 50.57\% & 49.91\% &  {\bf 47.64\%} & 0.56\% &  2.27\%  \\
Free~\citep{free19} & 47.19\% & 46.39\% &{\bf 44.20\%} & 0.80\% & 2.19\%  \\
YOPO~\citep{YOPO19} & 47.70\% & 47.07\% & {\bf 45.09\%} & 0.63\% & 1.98\% \\
 \bottomrule
\end{tabular}
\end{center}
\label{tab_recent}
\end{table*}

\paragraph{Results}
As shown in Table~\ref{tab_recent}, ODI-PGD uniformly outperforms na\"{i}ve-PGD against all six recently-proposed defense methods, lowering the estimated model accuracy by 1--17\%. In other words, ODI-PGD provides uniformly tighter upper bounds on the worst case model accuracy than na\"{i}ve-PGD. 
Additionally, The accuracy ranking of the defence methods for ODI-PGD is different from na\"{i}ve-PGD and $\text{PGD}_{1}$.
These results indicate that ODI-PGD might be {a better benchmark for comparing and evaluating different defense methods}, rather than na\"{i}ve-PGD and $\text{PGD}_{1}$.

\section{Additional results and experiments for ODS with black-box attacks}
\label{appendix:ap_black}

\subsection{Diversified samples by ODS}
\label{appendix:ap_black_diversity}
We empirically show that ODS can yield diversified changes in the output space of the target model, as shown in the right figures of Figure~\ref{figure1} of the main paper. 
Specifically, we evaluate the mean pairwise distance among outputs for different perturbations by ODS and 
compare it with the distance among outputs for random Gaussian sampling. 

We use pre-trained Resnet50~\citep{resnet16} and VGG19~\citep{VGG19} model 
as the target and surrogate models, respectively. 
We pick 100 images on ImageNet validation set and sample perturbations 10 times by each sampling method. For comparison, we normalize the perturbation to the same size in the input space. 
Then, the obtained pairwise distance on the target model by ODS is 0.79, which is 10 times larger than the pairwise distance by random Gaussian sampling (0.07). This indicates that the diversity by ODS is transferable. 

\subsection{Success rate curve in Section~\ref{sec_black_score} and Section~\ref{sec_black_decision} }
\label{appendix:ap_black_successcurve}
\begin{figure*}[htbp]
\centering
    \begin{tabular}{cc}
      \begin{minipage}{0.4\hsize}
        \begin{center}
          \includegraphics[width=0.9\textwidth]{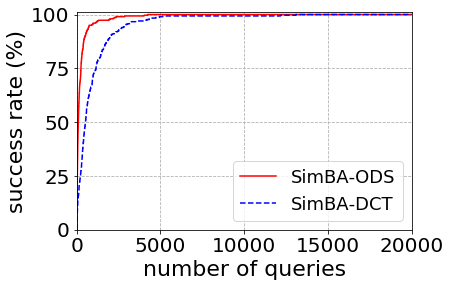}
        \end{center}
      \end{minipage}
      &
      \begin{minipage}{0.4\hsize}
        \begin{center}
          \includegraphics[width=0.9\textwidth]{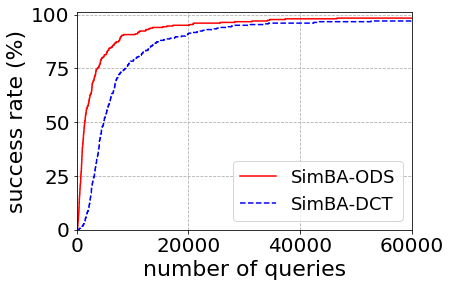}
        \end{center}
      \end{minipage}\\
      untargeted & targeted
    \end{tabular}
    \caption{Relationship between success rate and number of queries for score-based SimBA-ODS and SimBA-DCT.}
    \label{fig_append_simba}
\end{figure*}

\begin{figure*}[htbp]
\centering
    \begin{tabular}{cccc}
      \begin{minipage}{0.22\hsize}
        \begin{center}
          \includegraphics[width=1\textwidth]{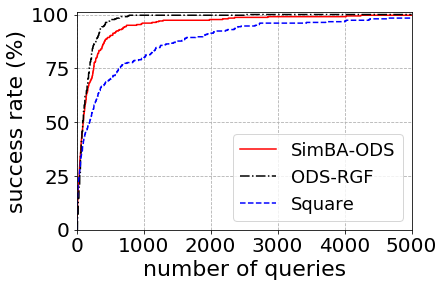}
        \end{center}
      \end{minipage}
      &
      \begin{minipage}{0.22\hsize}
        \begin{center}
          \includegraphics[width=1\textwidth]{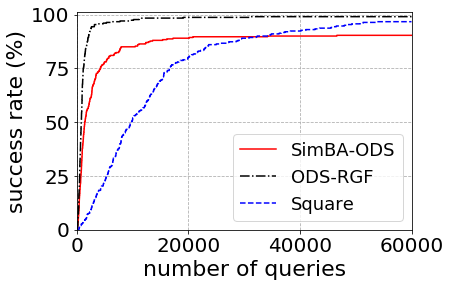}
        \end{center}
      \end{minipage} &
      \begin{minipage}{0.22\hsize}
        \begin{center}
          \includegraphics[width=1\textwidth]{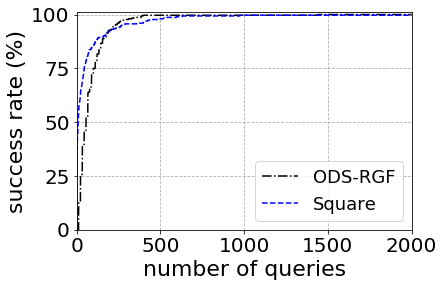}
        \end{center}
      \end{minipage}
      &
      \begin{minipage}{0.22\hsize}
        \begin{center}
          \includegraphics[width=1\textwidth]{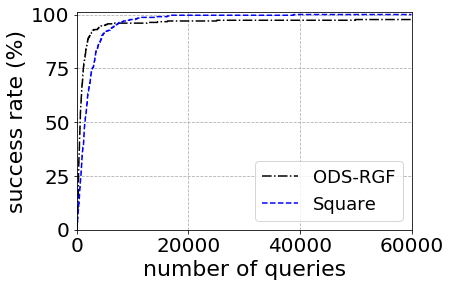}
        \end{center}
      \end{minipage}\\
      untargeted ($\ell_2$) & targeted  ($\ell_2$) & untargeted ($\ell_\infty$) & targeted  ($\ell_\infty$)
    \end{tabular}
    \caption{Relationship between success rate and number of queries for SimBA-ODS, ODS-RGF, and Square Attack. Each attack is evaluated with norm bound $\epsilon=5 (\ell_2), 0.05 (\ell_\infty)$.}
    \label{fig_append_square}
\end{figure*}

In Section~\ref{sec_black_score}, we demonstrated that SimBA-ODS outperformed state-of-the-art attacks in terms of the query-efficiency. As an additional result, we give the success rate curve of score-based attacks with respect to the number of queries in the experiments. Figure~\ref{fig_append_simba} shows how the success rate changes with the number of queries for SimBA-ODS and SimBA-DCT for the experiment of  Table~\ref{tab_black_score}. SimBA-ODS especially brings query-efficiency at small query levels. In Figure~\ref{fig_append_square}, we also describe the success rate curve for the experiment of Table~\ref{tab_black_square}. ODS-RGF outperforms other methods in the $\ell_2$ norm.

\begin{figure*}[htbp]
\centering
    \begin{tabular}{cc}
      \begin{minipage}{0.4\hsize}
        \begin{center}
          \includegraphics[width=0.9\textwidth]{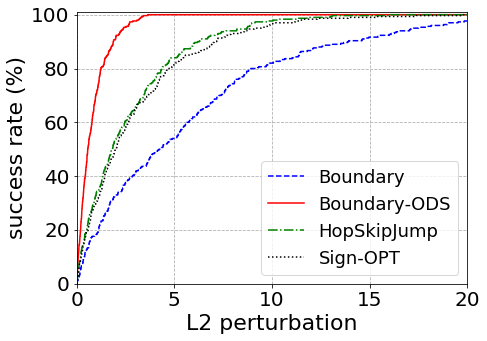}
        \end{center}
      \end{minipage}
      &
      \begin{minipage}{0.4\hsize}
        \begin{center}
          \includegraphics[width=0.9\textwidth]{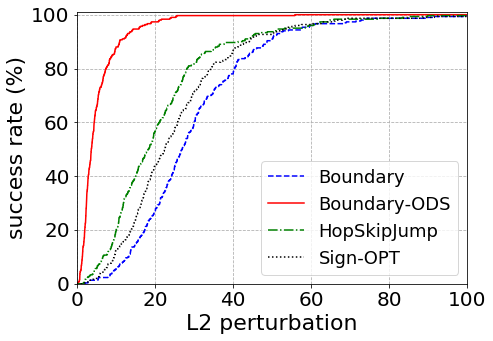}
        \end{center}
      \end{minipage}\\
      untargeted & targeted
    \end{tabular}
    \caption{Cumulative frequency distribution of $\ell_2$ perturbations at 10000 queries for decision-based attacks.}
    \label{fig_append_decision_curve}
\end{figure*}

In Section~\ref{sec_black_decision}, we demonstrated that Boundary-ODS outperformed state-of-the-art attacks in terms of median $\ell_2$ perturbation. Here, we depict the relationship between the success rate and perturbation size (i.e. the frequency distribution of the perturbations) to show the consistency of the improvement. 
Figure~\ref{fig_append_decision_curve} describes the cumulative frequency distribution of $\ell_2$ perturbations for each attack at 10000 queries.
Boundary-ODS consistently decreases $\ell_2$ perturbations compared to other attacks in both untargeted and targeted settings.

\subsection{Comparison of ODS with TREMBA}
\label{appendix:ap_black_TREMBA}
We run experiments to compare ODS with TREMBA, which is a state-of-the-art attack with surrogate models, as we mentioned in Section~\ref{sec_black_score_RGF}. 
TREMBA leverages surrogate models to learn a
low-dimensional embedding 
so as to obtain initial adversarial examples using a transfer-based attack and then update them using a score-based attack. 
Although TREMBA uses random sampling, ODS does not work well with TREMBA because random sampling of TREMBA is performed in the embedding space. In addition, it is difficult to directly compare attacks with ODS (e.g., ODS-RGF) and TREMBA because we do not discuss the combination of ODS with transfer-based attacks in this paper. 

However, we can start attacks with ODS (e.g., ODS-RGF) from images generated by any transfer-based attacks and compare the attack with TREMBA. 
We generate starting points by SI-NI-DIM~\citep{NesterovTransfer2020} (Scale-Invariant Nesterov Iterative FGSM integrated with diverse input method), which is a state-of-the-art transfer-based attack, and run ODS-RGF from these starting points. 

We adopt the same experiment setup as TREMBA~\citep{Huang20TREMBA}: we evaluate attacks against four target models (VGG19, ResNet34, DenseNet121, MobileNetV2) for 1000 images on ImageNet and use four surrogate models (VGG16, Resnet18, Squeezenet~\citep{squeezenet} and Googlenet~\citep{googlenet} ).
We set the same hyperparameters as in the original paper~\citep{Huang20TREMBA} for TREMBA. For fair comparisons, we set the same sample size (20) and use the same surrogate models as TREMBA for ODS-RGF. We also set step size of ODS-RGF as 0.001. 
As for SI-NI-DIM, we set hyperparameters referring to the paper~\citep{NesterovTransfer2020}: maximum iterations as 20, decay factor as 1, and number of scale copies as 5.
We describe results in Table~\ref{fig_black_TREMBA}. We can observe that 
ODS-RGF with SI-NI-DIM is comparable to TREMBA. 

We note that ODS is more flexible than TREMBA in some aspects.  First, TREMBA is specific in the $\ell_\infty$-norm, whereas ODS can be combined with attacks at least in $\ell_\infty$ and $\ell_2$-norms. 
In addition, TREMBA needs to train a generator per target class in targeted settings, whereas ODS does not need additional training.

\begin{table}[htb]
\caption{Comparison of ODS-RGF with TREMBA against four target models. The first two rows and bottom two rows describe results for untargeted (U) attacks and targeted (T) attacks, respectively. Targeted class for targeted attacks is class 0. }
\begin{center}
\setlength{\tabcolsep}{2pt}
\begin{tabular}{cc|cc|cc|cc|cc}
\toprule
&& \multicolumn{2}{c|}{VGG19}& \multicolumn{2}{c|}{ResNet34}& \multicolumn{2}{c|}{DenseNet121}& \multicolumn{2}{c}{MobilenetV2}\\
&attack& success & query& success & query& success & query& success & query\\ \hline
\multirow{2}{*}{U}&TREMBA~\citep{Huang20TREMBA}  & {\bf 100.0\%} & 34
&  {\bf 100.0\%} & 161 &  {\bf 100.0\%} & 157 &  {\bf 100.0\%}& 63\\
&SI-NI-DIM~\citep{NesterovTransfer2020} + ODS-RGF& {\bf 100.0\%} & {\bf 18}
&  99.9\% & {\bf 47} &   99.9\%  & {\bf 50} &  {\bf 100.0\%}& {\bf 29} \\ \hline
\multirow{2}{*}{T}&TREMBA~\citep{Huang20TREMBA} & {98.6\%} & 975
&  {96.7\%} & {\bf 1421} &  {\bf 98.5\%} & {\bf 1151} &  {\bf 99.0\%}& {\bf 1163}\\
&SI-NI-DIM~\citep{NesterovTransfer2020} + ODS-RGF& {\bf 99.4\%} & {\bf 634}
& {\bf 98.7\%} & {1578}   & {98.2\%} & 1550 & {98.3\%}& {2006} \\ 
\bottomrule
\end{tabular}
\end{center}
\label{fig_black_TREMBA}
\end{table}

\subsection{Performance of ODS against different target models}
\label{appendix:ap_black_targetchange}
In this paper, we used pre-trained ResNet50 model as the target model for all experiments in Section~\ref{sec_black_experiment}. Here we set pre-trained VGG19 model as the target model and run experiments to show that the efficiency of ODS is independent with target models.
As surrogate models, we replace VGG19 with ResNet50, i.e. we use four pre-trained models (ResNet50, ResNet34, DenseNet121, MobileNetV2).

We run experiments for SimBA-ODS in Section~\ref{sec_black_score} and Boundary-ODS in Section~\ref{sec_black_decision}. 
All settings except the target model and surrogate models are the same as the previous experiments.
In Table~\ref{tab_black_score_VGG} and \ref{tab_black_decision_VGG}, ODS significantly improves attacks against VGG19 model for both SimBA and Boundary Attack. This indicates that the efficiency of ODS does not depend on target models. 

\begin{table}[htb]
\caption{Query counts and $\ell_2$ perturbations for score-based Simple Black-box Attacks (SimBA) against pre-trained VGG19 model on ImageNet.}
\begin{center}
\setlength{\tabcolsep}{4pt}
\begin{tabular}{cc|ccc|ccc}
\toprule
&&  \multicolumn{3}{c}{untargeted} &\multicolumn{3}{|c}{targeted}  \\ \cline{3-8}
& num. of &success  & average & median $\ell_2$& success  & average & median $\ell_2$  \\ \
attack& surrogates& rate & query &  distance &  rate & query & distance  \\ \midrule
SimBA-DCT~\citep{Guo19} &0& {\bf 100.0\%} & 619 & 2.85 & {\bf100.0\%} & 4091 &6.81 \\
SimBA-ODS &4& {\bf 100.0\%} & {\bf 176} & {\bf 1.35} & 99.7\% & {\bf 1779 } & {\bf 3.31} \\
\bottomrule
\end{tabular}
\end{center}
\label{tab_black_score_VGG}
\end{table}

\begin{table}[htb]
\caption{Median $\ell_2$ perturbations for decision-based Boundary Attacks against pre-trained VGG19 model on ImageNet.}
\begin{center}
\begin{tabular}{cc|ccc|ccc}
\toprule
&&  \multicolumn{6}{c}{number of queries} \\
& num. of & \multicolumn{3}{c}{untargeted} &\multicolumn{3}{|c}{targeted} \\ %
attack&surrogates & 1000 & 5000 & 10000 & 1000 & 5000 & 10000 \\ \midrule
Boundary\citep{Brendel18} &0& 45.62&11.79&4.19 & 75.10&41.63&27.34 \\
Boundary-ODS&4 &  {\bf 6.03} & {\bf 0.69} & {\bf 0.43} & {\bf 24.11} & {\bf 5.44} & {\bf 2.97}\\
\bottomrule
\end{tabular}
\end{center}
\label{tab_black_decision_VGG}
\end{table}

\subsection{Effect of the choice of surrogate models}
\label{appendix:ap_black_surrogatechange}
In Section~\ref{sec_black_score} and \ref{sec_black_decision}, we mainly used four pre-trained models as surrogate models. To investigate the effect of the choice of surrogate models, we run attacks with seven different sets of surrogate models.
All settings except surrogate models are the same as the previous experiments.

Table~\ref{tab_black_score_surrogate} and \ref{tab_black_decision_surrogate} shows results for SimBA-ODS and Boundary-ODS, respectively. First, the first four rows in both tables are results for a single surrogate model. The degree of improvements depends on the model. ResNet34 gives the largest improvement and VGG19 gives the smallest improvement. Next, the fifth and sixth rows show results for sets of two surrogate models. By combining surrogate models, the query efficiency improves, especially for targeted attacks. This means that the diversity from multiple surrogate models is basically useful to make attacks strong. Finally, the performances in the seventh row are results for four surrogate models, which are not always better than results for the combination of two models (ResNet34 and DenseNet121). When the performances for each surrogate model are widely different, the combination of those surrogate models could be harmful.

\begin{table}[htb]
\caption{Query counts and $\ell_2$ perturbations for SimBA-ODS attacks with various sets of surrogate models. In the column of surrogate models, R:ResNet34, D:DenseNet121, V:VGG19, M:MobileNetV2.}
\begin{center}
\begin{tabular}{cc|ccc|ccc}
\toprule
&& \multicolumn{3}{c}{untargeted} &\multicolumn{3}{|c}{targeted}  \\ \cline{3-8}
surrogate& & success  & average & median $\ell_2$& success  & average & median $\ell_2$  \\ \
models&num.&  rate & query &  distance &  rate & query & distance  \\
\midrule
R&1& {\bf 100.0\%}  & {274} & 1.35 & 95.3\%& 5115 & 3.50  \\
D&1& {\bf 100.0\%}  &342&1.38 &96.7\% & 5282 & 3.51\\
V&1& {\bf 100.0\%}  &660 &1.78 &88.0\% & 9769 & 4.80 \\
M&1& {\bf 100.0\%}  &475&1.70 &95.3\%& 6539 & 4.53\\
R,D&2& {\bf 100.0\%} &{\bf 223}&{\bf 1.31} & 98.0\%& {\bf 3381} & {\bf 3.39}\\
V,M&2& {\bf 100.0\%} &374&1.60 & 96.3\%& 4696 & 4.27\\
R,V,D,M &4&  {\bf 100.0\%}& {241} &  1.40 & {\bf 98.3\%} & {3502} & {3.55}\\
\bottomrule
\end{tabular}
\end{center}
\label{tab_black_score_surrogate}
\end{table}

\begin{table}[htb]
\caption{Median $\ell_2$ perturbations for Boundary-ODS attacks with various sets of surrogate models. In the column of surrogate models, R:ResNet34, D:DenseNet121, V:VGG19, M:MobileNetV2.}
\begin{center}
\begin{tabular}{cc|ccc|ccc}
\toprule
& & \multicolumn{6}{c}{number of queries} \\ 
surrogate& & \multicolumn{3}{c}{untargeted} &\multicolumn{3}{|c}{targeted} \\ 
models & num. & 1000 & 5000 & 10000 & 1000 & 5000 & 10000 \\ \midrule
R&1 & 9.90&1.41&0.79 & 31.32&11.49&7.89\\
D&1 & 10.12&1.39&0.76 &32.63&11.30&7.44\\
V&1 & 22.68&3.47&1.52 & 49.18&24.26&17.75 \\
M&1 & 20.67&2.34&1.10 & 44.90&18.62&12.01\\
R,D&2 & {\bf 7.53}&1.07&0.61 & {\bf 26.00} & 8.08& 6.22\\
V,M&2 & 17.60& 1.70& 0.92&39.63&14.97&9.21 \\
R,V,D,M &4 &   7.57 & {\bf 0.98} & {\bf 0.57} &  27.24 & {\bf 6.84} & {\bf 3.76}\\
\bottomrule
\end{tabular}
\end{center}
\label{tab_black_decision_surrogate}
\end{table}

In Section~\ref{sec_black_score_RGF}, we compared ODS-RGF with P-RGF only using the ResNet34 surrogate model. To show the effectiveness of ODS-RGF is robust to the choice of surrogate models, we evaluate ODS-RGF with different surrogate models. Table~\ref{fig_black_RGF_VGG} shows the query-efficiency of ODS-RGF and P-RGF with the VGG19 surrogate model. We can observe that ODS-RGF outperforms P-RGF for all settings 
and the results are consistent with the experiment in Section~\ref{sec_black_score_RGF}.

\begin{table}[htb]
\caption{Comparison between ODS-RGF and P-RGF with the VGG19 surrogate model. Settings in the comparison are the same as Figure~\ref{fig_black_RGF}. }
\begin{center}
\setlength{\tabcolsep}{3.7pt}
\begin{tabular}{ccc|ccc|ccc}
\toprule
& & &\multicolumn{3}{c}{untargeted} &\multicolumn{3}{|c}{targeted}  \\ \cline{4-9}
  &  &num. of & & average & median $\ell_2$ & success  &  average & median $\ell_2$   \\ 
norm& attack &surrogates & success  & queries & perturbation & success  &  queries & perturbation  \\ 
\midrule
\multirow{3}{*}{$\ell_2$}&RGF & 0& \textbf{100.0\%} & 633 & 3.07 &\textbf{99.3\%} &3141 & 8.23\\
&P-RGF~[25] &1& \textbf{100.0\%} &467 & 3.03
&{97.0\%} &3130 & 8.18\\
&ODS-RGF& 1& \textbf{100.0\%} &\textbf{294}& \textbf{2.24}
&{98.0\%} &\textbf{2274} & \textbf{6.60}\\ \hline
\multirow{3}{*}{$\ell_\infty$} &RGF  &0& {97.0\%} & 520 & - 
&{25.0\%} & 2971   & -  \\
&P-RGF~[25]  &1& {98.7\%} &337& - & {29.0\%} & 2990 & -\\
&ODS-RGF& 1& \textbf{99.7\%} &\textbf{256}& - & \textbf{45.7\%} & \textbf{2116} & -  \\
\bottomrule
\end{tabular}
\end{center}
\label{fig_black_RGF_VGG}
\end{table}

\subsection{Effect of the number of surrogate models for the experiment in Section~\ref{sec_black_limited}}
\label{appendix:ap_black_OOD_surrogatenum}
We described that surrogate models with limited out-of-distribution training dataset are still useful for ODS in Section~\ref{sec_black_limited}. 
In the experiment, we used five surrogate models with the same ResNet18 architecture.  
Here, we reveal the importance of the number of surrogate models through experiments with the different number of models. 
Table~\ref{tab_black_limited_append} shows the result for Boundary-ODS with the different number of surrogate models. With the increase of the number of models, the query efficiency consistently improves.

\begin{table}[htb]
\caption{Median $\ell_2$ perturbations for Boundary-ODS attacks with different number of surrogate models against out-of-distribution images on ImageNet.}
\begin{center}
\begin{tabular}{c|ccc|ccc}
\toprule
& \multicolumn{6}{c}{number of queries} \\ 
num. of&  \multicolumn{3}{c}{untargeted} &\multicolumn{3}{|c}{targeted} \\ 
surrogates & 1000 & 5000 & 10000 & 1000 & 5000 & 10000 \\ \midrule
1  & 19.45 & 2.90 & 1.66 & 47.86& 25.30 & 20.46\\ 
2  & 15.45 & 2.42 & 1.35 & 43.45 & 19.30 & 13.78\\ 
3  & 13.75& 1.96&1.14& {\bf 41.63}& 16.91 & 11.14\\ 
4  & 14.23& 1.86 & 1.21 & 41.65& 14.86 & 9.64\\ 
5  & {\bf 11.27}& {\bf 1.63}& {\bf 0.98}& 41.67& {\bf 13.72} & {\bf 8.39}\\ 
\bottomrule
\end{tabular}
\end{center}
\label{tab_black_limited_append}
\end{table}

\subsection{Score-based attacks with ODS against out-of-distribution images}
\label{appendix:ap_black_OOD_scorebased}
In Section~\ref{sec_black_limited}, we demonstrated that the decision-based Boundary-ODS attack works well even if we only have surrogate models trained with limited out-of-distribution dataset.
Here, we evaluate score-based SimBA-ODS with these surrogate models. Except surrogate models, we adopt the same setting as Section~\ref{sec_black_score}.

In Table~\ref{tab_black_OOD_SimBA}, SimBA-ODS with out-of-distribution dataset outperforms SimBA-DCT in untargeted settings. 
In targeted settings, while SimBA-ODS improves the $\ell_2$ perturbation, the average queries for SimBA-ODS are comparable with SimBA-DCT. 
We hypothesize that it is because ODS only explores the subspace of the input space. The restriction to the subspace may lead to bad local optima. 
We can mitigate this local optima problem by applying random sampling temporally when SimBA-ODS fails to update a target image in many steps in a low.

We note that decision-based Boundary-ODS with OOD dataset is effective, as shown in Section~\ref{sec_black_limited}. 
We hypothesize that the difference in effectiveness is because Boundary-ODS does not use scores of the target model and thus does not trap in local optima.

\begin{table}[htb]
\caption{Query counts and $\ell_2$ perturbations for SimBA-ODS attacks with surrogate models trained with OOD images on ImageNet.}
\begin{center}
\setlength{\tabcolsep}{4pt}
\begin{tabular}{c|ccc|ccc}
\toprule
& \multicolumn{3}{c}{untargeted} &\multicolumn{3}{|c}{targeted}  \\ \cline{2-7}
& success  & average & median $\ell_2$ & success  & average & median $\ell_2$  \\ \
attack&  rate & queries & perturbation &  rate & queries &  perturbation  \\ \midrule
SimBA-DCT~\citep{Guo19} & {\bf 100.0\%} & 909 & 2.95 & {\bf 97.0\%} & 7114 &7.00 \\
SimBA-ODS 
(OOD dataset) & {\bf 100.0\%} & {\bf 491} & {\bf 1.94} & 94.7\% & {\bf 6925} & {\bf 4.92} \\ \hline
\begin{tabular}{c}
SimBA-ODS (full dataset) 
\end{tabular} & {100.0\%} & {242} & {1.40} & {98.3\%} & {3503} & {3.55} \\
\bottomrule
\end{tabular}
\end{center}
\label{tab_black_OOD_SimBA}
\end{table}

\subsection{ODS against robust defense models}
\label{appendix:ap_black_defensemodel}
In this paper, we mainly discuss transferability when surrogate models and the target model are trained with similar training schemes. On the other hand, it is known that transferability decreases when models are trained with different training schemes, 
e.g. the target model uses adversarial training and surrogate models use natural training. 
If all surrogate models are trained with natural training scheme, 
ODS will also not work against adversarially trained target models. 
However, we can mitigate the problem by simultaneously using surrogates obtained with various training schemes (which are mostly publicly available). In order to confirm this, we run an experiment to attack a robust target model using SimBA-ODS with both natural and robust surrogate models (a natural model and a robust model). In Table~\ref{tab_black_advmodel}, the first row shows the attack performance of SimBA-DCT (without surrogate models) and the others show the performance of SimBA-ODS. 
In the fourth row of Table~\ref{tab_black_advmodel}, SimBA-ODS with natural and robust surrogate models significantly outperforms SimBA-DCT without surrogate models. This suggests that if the set of surrogates includes one that is similar to the target, ODS still works (even when some other surrogates are "wrong").
While the performance with natural and robust surrogate models slightly underperforms single adversarial surrogate model in the third row, dynamic selection of surrogate models during the attack will improve the performance, as we mentioned in the conclusion of the paper.

\begin{table*}[htbp]
\caption{Transferability of ODS when training schemes of surrogate models are different from the target model. 
R50 shows pretrained ResNet50 model, and R101(adv) and R152(adv) are adversarially trained ResNeXt101 and ResNet152 denoise models from~\citep{featureDenoise19}, respectively. All attacks are held in the same setting as Section~\ref{sec_black_score}. 
}
\begin{center}
\begin{tabular}{cc|ccc}
\toprule
target & surrogate & success rate  & average queries & median $\ell_2$ perturbation \\ 
 \midrule
R101(adv) &- &  89.0\% & 2824& 6.38\\
R101(adv) &R50 & 80.0\% & 4337& 10.15\\
R101(adv) &R152(adv) & \textbf{98.0\%} &\textbf{1066} & \textbf{4.93}\\ 
R101(adv) &R50, R152(adv) & \textbf{98.0\%} &1304 & {5.62}\\
\bottomrule         
\end{tabular}
\end{center}
\label{tab_black_advmodel}
\end{table*}

\section{Relationship and Comparison between ODS and MultiTargeted}
\label{appendix:ap_multitargeted}
In this section, we describe that ODS gives better diversity than the MultiTargeted attack~\citep{MT19} for initialization and sampling. 

MultiTargeted is a variant of white-box PGD attacks, which maximizes $f_t(\bmx)-f_y(\bmx)$ where $\bmf(\bmx)$ is logits, $y$ is the original label and $t$ is a target label. The target label is changed per restarts. In other words, MultiTargeted moves a target image to a particular direction in the output space, which is represented as like $\bmw_{\text{d}}=(1,0,-1,0)$ where 1 and -1 correspond to the target and original label, respectively. Namely, the procedure of MultiTargeted is technically similar to ODS.

However, there are some key differences between MultiTargeted and ODS. One of the difference is the motivation. MultiTargeted was proposed as a white-box attack and the study only focused on $\ell_p$-bounded white-box attacks. On the other hand, our study gives broader application for white- and black-box attacks. As far as we know, ODS is the first method which exploits the output diversity for initialization and sampling.

Another difference is the necessity of the original label of target images. 
ODS does not require the original class of the target image,
and thus ODS is applicable for black-box attacks 
even if surrogate models are trained with out-of-distribution training dataset, as shown in Section~\ref{sec_black_limited}.
On the other hand, 
since MultiTargeted exploits the original label of target images to calculate the direction of the attack, 
we cannot apply MultiTargeted to sampling for black-box attacks against out-of-distribution images.

Finally, the level of diversity is also different. 
As we mentioned in Section~\ref{sec_related}, the direction of MultiTargeted is restricted to away from the original class.
This restriction could be harmful for diversity because the subspace to explore directions is limited. 
To show this statement, we apply MultiTargeted to initialization for white-box attacks and sampling for black-box attacks, and demonstrate that ODI provides better diversity than MultiTargeted for initialization and sampling (especially for sampling).

\paragraph{Initialization in white-box settings}
We apply MultiTargeted to initalization for white-box attacks in Section~\ref{sec_white_various}. 
Table~\ref{tab_appendix_MTwhite} represents the comparison of the attack performance with initialization by MultiTargeted and ODI.
For PGD attacks, MultiTargeted is slightly better than ODI. We hypotheses that it is because MultiTargeted was developed as a variant of PGD attacks and the initialization by MultiTargeted also works as an attack method. 
On the other hand, ODI outperforms MultiTargeted for C\&W attacks. In this setting, MultiTargeted does not work as an attack method, and thus the difference in the diversity makes the difference in the performance.
\begin{table*}[htb]
\caption{Comparison of model performance under attacks with MultiTargeted (MT) and ODI. The values are model accuracy (lower is better) for PGD and the average of the minimum $\ell_2$ perturbations (lower is better) for C\&W. All results are the average of three trials. Results for ODI are from Table~\ref{tab_Linf}.
}
\begin{center}
\setlength{\tabcolsep}{4.5pt}
\begin{tabular}{c|cc|cc}
\toprule
 & \multicolumn{2}{|c}{PGD } &
\multicolumn{2}{|c}{C\&W} \\ 
model &  MT & ODI & MT & ODI 
\\ \midrule
MNIST & $\textbf{89.95}\pm 0.05\%$ & ${90.21}\pm 0.05\%$ &  ${2.26}\pm0.01$ &  $\textbf{2.25}\pm0.01$  \\ 
CIFAR-10 &  $\textbf{44.33}\pm 0.01\%$ &${44.45}\pm 0.02\%$ & $0.69\pm0.01$  &$\textbf{0.67}\pm0.00$ \\ 
ImageNet  &  $\textbf{42.2}\pm 0.0\%$ &  ${42.3}\pm 0.0\%$& $2.30\pm0.01$ & $\textbf{1.32}\pm0.01$ \\ 
\bottomrule         
\end{tabular}
\end{center}
\label{tab_appendix_MTwhite}
\end{table*}

\begin{figure}[ht]
\centering

    \begin{tabular}{cc}
      \begin{minipage}{0.35\hsize}
        \begin{center}
          \includegraphics[width=1.0\textwidth]{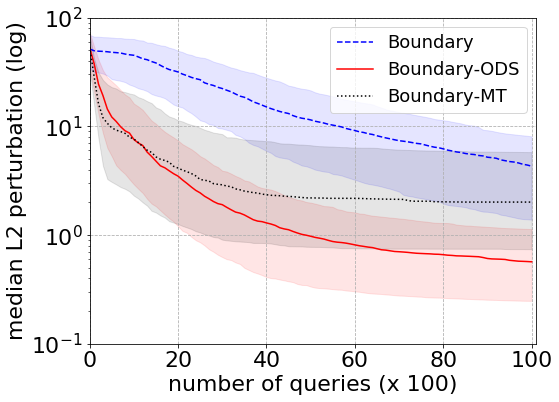}
        \end{center}
      \end{minipage}
        &
      \begin{minipage}{0.35\hsize}
        \begin{center}
          \includegraphics[width=1.0\textwidth]{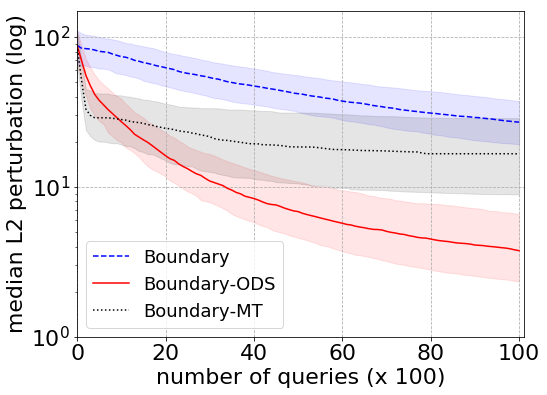}
        \end{center}
      \end{minipage}
      \\
      Untargeted & Targeted 
    \end{tabular}
    \caption{Relationship between median $\ell_2$ perturbations and the number of queries for Boundary Attack with ODS and MultiTargeted. Error bars show 25\%ile and 75\%ile of $\ell_2$ perturbations. }
    \label{fig_append_MTblack}
\end{figure}

\paragraph{Sampling in black-box settings} 
We use MultiTargeted for sampling on the Boundary Attack in Section~\ref{sec_black_decision} (called Boundary-MT), and compare it with Boundary-ODS. Table~\ref{tab_append_MTblack} and Figure~\ref{fig_append_MTblack} show the results of the comparison. While Boundary-MT outperforms the original Boundary Attack, Boundary-ODS finds much smaller adversarial perturbation than Boundary-MT. 

In Figure~\ref{fig_append_MTblack}, Boundary-MT slightly outperforms Boundary-ODS at small queries. We hypotheses that it is because MultiTargeted not works for providing diversity, but works for the approximation of gradients of the loss function. However, with the number of queries, the curve of Boundary-MT is saturated, and Boundary-MT underperforms Boundary-ODS. This is an evidence that the restriction of directions is harmful for sampling. %

\begin{table}[htb]
\caption{Median $\ell_2$ perturbations for Boundary Attack with ODS and MultiTargeted (MT).}
\begin{center}
\begin{tabular}{c|ccc|ccc}
\toprule
&  \multicolumn{6}{c}{number of queries} \\
&  \multicolumn{3}{c}{untargeted} &\multicolumn{3}{|c}{targeted} \\ 
attack & 1000 & 5000 & 10000 & 1000 & 5000 & 10000 \\ \midrule
Boundary~\citep{Brendel18} & 45.07& 11.46 & 4.30 & 73.94 &41.88 &27.05  \\
Boundary-ODS &  {\bf 7.57} & {\bf 0.98} & {\bf 0.57} & {\bf 27.24} & {\bf 6.84} & {\bf 3.76}\\
Boundary-MT & 7.65&2.20&2.01&28.16&18.48&16.59  \\
\bottomrule
\end{tabular}
\end{center}
\label{tab_append_MTblack}
\end{table}

\end{document}